\documentclass{article} %
\usepackage{iclr2027_conference,times}

\usepackage{amsmath,amsfonts,bm}

\def\eqref#1{equation~\ref{#1}}

\def\1{\bm{1}}

\DeclareMathAlphabet{\mathsfit}{\encodingdefault}{\sfdefault}{m}{sl}
\SetMathAlphabet{\mathsfit}{bold}{\encodingdefault}{\sfdefault}{bx}{n}

\newcommand{\Ours}{SiMDex\xspace}
\newcommand{\Baseline}{GR-Dexter\xspace}

\usepackage{minitoc}

\usepackage[utf8]{inputenc}
\usepackage[T1]{fontenc}

\usepackage{amsmath, amssymb, amsfonts, mathtools}
\usepackage{bbm}                     %
\usepackage{bm}                      %

\usepackage{booktabs}                %
\usepackage{array}                   %
\usepackage{multirow}                %
\usepackage{makecell}                %
\usepackage{colortbl}                %
\usepackage{tabulary}                %
\usepackage{tabularx}                %
\usepackage{threeparttable}          %
\usepackage{arydshln}                %
\usepackage{adjustbox}               %
\usepackage{siunitx}                 %

\usepackage{graphicx}
\usepackage{caption}
\usepackage{subcaption}
\usepackage{float}                   %
\usepackage{wrapfig}                 %
\usepackage{overpic}                 %

\usepackage{xcolor}
\definecolor{mygray}{HTML}{f0f0f0}
\definecolor{mygreen}{HTML}{35cd2d}
\definecolor{COLOR_MEAN}{HTML}{f0f0f0}
\definecolor{GREEN}{HTML}{0aa344}
\definecolor{dgreen}{rgb}{0.0,0.6,0.0}
\definecolor{gainpos}{HTML}{2C7A6B}  %
\definecolor{gainneg}{HTML}{C0392B}  %

\usepackage{pifont}                  %
\usepackage{bbding}                  %
\usepackage[pro]{fontawesome5}       %

\newcommand{\dpos}[1]{\textcolor{gainpos}{#1}}  %
\newcommand{\dneg}[1]{\textcolor{gainneg}{#1}}  %

\usepackage{tikz}

\usepackage[ruled,lined,linesnumbered]{algorithm2e}  %
 
\usepackage{listings}
\usepackage{verbatim}

\usepackage{hyperref}
\usepackage{url}

\usepackage{enumitem}                %
\usepackage{xspace}                  %
\usepackage{orcidlink}               %

\newcommand{\myparagraph}[1]{\vspace{0.1em}\noindent\textbf{#1}}

\title{{\frenchspacing\spaceskip=\fontdimen2\font plus \fontdimen3\font minus 1.2\fontdimen4\font
SiMDex: Mining Similar Egocentric Videos for Cross-Embodiment Dexterous Manipulation}}

\author{%
Nie Lin$^{1}$ \quad Takehiko Ohkawa$^{1}$ \quad Sijin Chen$^{2,3}$ \quad Ruoshi Wen$^{2}$ \quad Zhuohang Li$^{4}$ \\
\textbf{Liqun Huang$^{2}$ \quad Zhengming Zhu$^{2}$ \quad Yiming Bao$^{5}$ \quad Yunfei Li$^{2}$ \quad Minjie Cai$^{1}$} \\
\textbf{Xiao Ma$^{2}$ \quad Wei Xu$^{2,\dagger}$ \quad Yoichi Sato$^{1,\dagger}$} \\[0.6ex]
$^{1}$The University of Tokyo \quad $^{2}$ByteDance Seed \quad $^{3}$The University of Hong Kong \\
$^{4}$Shanghai Jiao Tong University \quad $^{5}$Tsinghua University
}

\iclrfinalcopy %
\begin{document}

\maketitle
\lhead{Preprint} %

\begingroup
\renewcommand{\thefootnote}{\fnsymbol{footnote}}
\footnotetext[2]{Corresponding authors}
\endgroup

\begin{abstract}
Recent years have witnessed an explosive trend of scaling ego-centric human videos for robot manipulation, yet it remains unclear which data actually benefits dexterous manipulation.
We present \textbf{\Ours}, a \underline{\textbf{si}}milarity-based data \underline{\textbf{m}}ining framework that casts human data selection for VLA post-training in \underline{\textbf{dex}}terous manipulation as a recommendation problem.
For each robot demonstration, \Ours runs a recall--ranking--re-ranking cascade over $\sim$32M egocentric human samples: two stages, recall and ranking, select a task-relevant subset in a morphology-agnostic action space, and re-ranking verifies it on pixels, with no change to VLA architecture or training.
Against a strong baseline trained with an equal amount of randomly sampled human data, SiMDex uses only $\sim$1.49M mined samples ($<$5\% of the pool) yet improves the overall success rate from 47.7\% to 61.7\%---showing that selective curation outperforms indiscriminate data mixing.
\end{abstract}
\vspace{-1.5ex}
\begin{center}
\textbf{Project page:} \url{https://lin-nie.github.io/SiMDex/}
\end{center}
\section{Introduction}
\label{sec:introduction}
Vision-Language-Action (VLA) models have driven strong progress in robotic manipulation, with performance scaling predictably as training data grows~\citep{brohan:rss23, brohan:arxiv23, black:rss25, intelligence:corl25, kim:corl24}.
Yet scaling robot data is itself the bottleneck: the high cost, poor portability, and teleoperation expertise of robot hardware drive researchers to seek other ways to scale up manipulation data~\citep{oxe:icra24, khazatsky:rss24, ohkawa:arxiv2026}.
Among various approaches, ego-centric human manipulation data has emerged to be a promising data source for scaling up generalist robot policies.
Large-scale egocentric datasets~\citep{grauman:cvpr22, damen:ijcv22, li2026egolive, punamiya:rss26} already span thousands of hours and keep growing as wearable devices spread and embodied AI drives demand.

From thousands of hours~\citep{grauman:cvpr22} to tens of thousands of hours~\citep{nvidia:arxiv26} to a million hours~\citep{deng:arxiv25}, the vast expansion of human ego-centric manipulation data now casts new questions: \textbf{how can we use human ego-centric data more efficiently?}
In practice, egocentric pools are extraordinarily heterogeneous—spanning cooking, socializing, sports, and countless other activities—of which only a small fraction is relevant to any given manipulation task.
Naively training on all of it indiscriminately injects noise and dilutes task-relevant supervision, while discarding human data entirely wastes the cross-embodiment knowledge acquired during large-scale pre-training and the generalization it brings to real-world robot deployment.

In large-scale vision and language modeling, targeted data curation has proven essential to unlocking the potential of web-scale corpora~\citep{gadre:neurips23, abbas:arxiv23, penedo:neurips24}. Carefully selected subsets often match or surpass models trained on far larger unfiltered data; robot learning from human video now stands at the same crossroads.
Once a VLA has been pre-trained on large-scale egocentric videos, \emph{which} human data to revisit for the target task during post-training matters more than \emph{how much}.
The key is not to scale indiscriminately, but to select precisely.

Data curation has received growing attention in robot learning, yet existing methods operate at a different scope from ours.
One line filters or reweights demonstrations within robot datasets—e.g., CUPID~\citep{agia:corl25} and Re-Mix~\citep{hejna:corl24}—or retrieves behaviors from offline robot data, all operating entirely within robot data rather than mining an external human pool.
Closest to ours, recent work retrieves task-relevant human videos to aid policy learning~\citep{zhu:iros25, papagiannis:icra25}, but operates over small video banks and feeds retrieved cues as auxiliary inputs or in-context exemplars, requiring changes to the policy architecture.
What remains unaddressed is task-aware, per-demonstration retrieval at the scale of tens of millions of cross-embodiment samples that, based on fine-grained action-level similarity, directly augments VLA training without modifying the policy architecture or the training procedure.

We present \Ours, a similarity-based data mining framework that casts human data selection for VLA post-training as a recommendation problem: as in industrial recommenders that retrieve from billion-scale catalogs, the tens-of-millions-sample human pool makes exhaustive comparison infeasible.
\Ours adopts a recall--ranking--re-ranking cascade that progressively distills task-relevant subsets for each robot demonstration.
Its two selection stages, recall and ranking, produce the mined subset, while re-ranking audits that subset on pixels, with a signal that shares no information with the two action-space stages.
The cascade operates in a morphology-agnostic action space that requires no architectural change---shaping \emph{what} a VLA learns from, not \emph{how}.
Notably, \Ours re-mines the same egocentric corpus used for pre-training, revisiting it task-aware rather than collecting new data.
Large-scale egocentric collection pays off twice, once in breadth and once in precision.

Across three real-world dexterous manipulation tasks, SiMDex improves the overall success rate from 47.7\% to 61.7\% using fewer than 5\% of the available human samples, and it leads the baseline at every robot-data budget we test, from a quarter to twice the default amount.

In summary, our main contributions are threefold:
\begin{itemize}
    \item \textbf{Recommendation view.} We reframe human data selection for VLA post-training as a recommendation problem, re-mining the \emph{same} egocentric corpus used for pre-training—now task-aware—so that large-scale collection pays off twice: in breadth, then in precision.
    \item \textbf{Simple, scalable mining pipeline.} We instantiate this view as a cascade: two selection stages, recall and ranking, mine a training subset from a large human pool in a morphology-agnostic action space, and re-ranking verifies it on pixels, with no change to the VLA.
    \item \textbf{Empirical gains.} On real-world dexterous manipulation tasks, \Ours substantially improves overall success over an identical baseline trained on equal randomly-sampled human data, and the gain holds at each of the four robot-data budgets we test.
\end{itemize}

\section{Related Works}
\label{sec:related_works}

\myparagraph{VLA Models for Robotic Manipulation.}
Recent VLA models achieve strong instruction-conditioned manipulation through large-scale pre-training~\citep{brohan:rss23, brohan:arxiv23, team:rss24, kim:corl24, chi:ijrr24, li:arxiv24_cogact, black:rss25, intelligence:corl25, intelligence:rss26, cheang:arxiv25_gr3}, yet most remain limited to low-DoF grippers. Extending them to high-DoF dexterous hands is far harder due to the expanded action space and contact-rich dynamics~\citep{wang:rss24, handa:icra20}. Recent efforts use hierarchical planning--control~\citep{zhong:aaai26, yuan:arxiv25} or end-to-end prediction~\citep{wen:arxiv25, luo:arxiv25}, but both are bottlenecked by the scarcity of real dexterous data, as teleoperation~\citep{wang:rss24, wen:arxiv25_teleop} yields datasets orders of magnitude smaller than gripper counterparts. Rather than scaling robot-data collection, \Ours takes a complementary, model-agnostic approach: mining task-relevant human egocentric data to augment VLA training without architectural change.

\myparagraph{Learning from Human Videos.}
Human videos have been leveraged for robot learning, from visual representation pre-training~\citep{nair:icml22, ma:icml23, majumdar:icra23} to direct policy learning~\citep{bahl:rss22, wang:corl23, kareer:icra25} and VLA pre-training from egocentric videos~\citep{yang2025egovla, li2025scalable}, with another line bridging the embodiment gap via domain-adaptive imitation~\citep{punamiya:neurips25}, data editing~\citep{lepert:icra26}, robot-data generation~\citep{mu:rss26}, or shared action-space design~\citep{chen:arxiv26}. Large-scale benchmarks such as Ego4D~\citep{grauman:cvpr22}, Ego-Exo4D~\citep{grauman:cvpr24}, and EgoDex~\citep{hoque:iclr26} provide rich foundations, and \citet{nvidia:arxiv26} show a log-linear scaling law between human data scale and validation loss---though their mid-training requires aligned human--robot play under the robot's task distribution, making data--task correspondence a manual prerequisite. More broadly, these methods consume human data at the dataset level without tailoring selection to the downstream task~\citep{nair:icml22, radosavovic:corl23, nvidia:arxiv26, luo:arxiv25, kareer:icra25}, inheriting noise from irrelevant activities. \Ours instead performs task-aware hierarchical retrieval to identify the most relevant human demonstrations for VLA training.

\myparagraph{Data Curation for Robot Learning.}
Data curation is central to foundation models: in vision and language, filtered or deduplicated subsets match or surpass models trained on larger corpora~\citep{gadre:neurips23, abbas:arxiv23, tirumala:neurips23}, and similar-sample mining benefits representation pre-training~\citep{nie:iclr25}. Within robot learning, curation has so far stayed confined to robot data---estimating demonstration influence~\citep{agia:corl25}, optimizing mixing weights~\citep{hejna:corl24}, or retrieving behaviors from robot datasets~\citep{du:rss23}. A few methods do retrieve human videos~\citep{zhu:iros25, papagiannis:icra25}, but use them as auxiliary or in-context inputs over limited collections. To our knowledge, no prior approach offers task-aware, per-demonstration retrieval that mines a massive cross-embodiment pool by fine-grained action-level similarity and feeds the result directly into VLA training without touching the architecture. \Ours fills this gap by framing cross-embodiment selection as a retrieval problem: 
like industrial recommenders over billion-scale catalogs, we adapt a recall--ranking--re-ranking cascade to mine a task-relevant subset from tens of millions of samples. Re-ranking verifies retrieval quality on pixels via a motion signal shared by both embodiments, independent of the action space.
\section{Method}
\label{sec:method}
We propose \Ours, a similarity-based data mining framework that retrieves high-quality human data from large-scale egocentric video to enhance VLA training for dexterous manipulation.
Operating over a pool of tens of millions of human samples, \Ours deliberately adopts a lightweight, scalable similarity cascade rather than a heavy learned cross-embodiment alignment module: at this scale, a fast and reliable retrieval pipeline is what makes task-aware mining practical.

\subsection{Unified Morphology-Agnostic Representation}
\label{sec:representation}
A prerequisite for cross-embodiment mining is a representation that abstracts away the morphological gap between robot and human hands.
We adopt a fingertip-centric geometric representation that captures task-relevant contact structure while remaining agnostic to joint-level kinematics, and map both sources onto it through one shared wrist-local retargeting procedure.

\paragraph{Wrist-local retargeting.}
For each source, we recover per-frame wrist poses and fingertip positions in 3D.
Letting $\bm{T}_t \in \mathbb{R}^{4\times 4}$ be the wrist pose at time $t$, with rotation $\bm{R}_t \in \mathbb{R}^{3\times 3}$ and translation $\bm{o}_t \in \mathbb{R}^{3}$, and $\bm{q}_t \in \mathbb{R}^{3}$ a fingertip position in world coordinates, we re-express each fingertip relative to its wrist as $\bm{q}^{\mathrm{loc}}_t = \bm{R}_t^{\top}(\bm{q}_t - \bm{o}_t) \in \mathbb{R}^{3}$, isolating intrinsic grasp geometry from workspace location.
The five wrist-local fingertips of one hand form a fingertip configuration $\bm{p}_t \in \mathbb{R}^{15}$.
The wrist action $\bm{d}_t \in \mathbb{R}^{6}$ gives a future wrist pose in the current wrist frame, with translational part $\bm{R}_t^{\top}(\bm{o}_{t+k} - \bm{o}_t) \in \mathbb{R}^{3}$ and rotational part the axis-angle vector of $\bm{R}_t^\top \bm{R}_{t+k}$ for the $k$-th future step.
Together, $(\bm{d}_t, \bm{p}_t)$ form the shared descriptor onto which both embodiments are mapped.
We write $\bm{p}_t$ for the observed configuration and $\bm{f}_t \coloneqq \bm{p}_{t+k}$ for the one predicted at the same future step.

\paragraph{Unified action.}
After preprocessing, both sources converge to the same per-timestep description.
Its embodiment-shared part concatenates bimanual wrist actions $\bm{a}^{\mathrm{wr}}_t = (\bm{d}^{\mathrm{L}}_t, \bm{d}^{\mathrm{R}}_t) \in \mathbb{R}^{12}$ and bimanual fingertip targets $\bm{a}^{\mathrm{ft}}_t = (\bm{f}^{\mathrm{L}}_t, \bm{f}^{\mathrm{R}}_t) \in \mathbb{R}^{30}$, 42 embodiment-shared dimensions in total.
Because both are expressed relative to the current wrist, these dimensions are invariant to workspace location and require no matching of kinematic structures across embodiments.
They are the basis for both mining (Sec.~\ref{sec:mining}) and supervision (Sec.~\ref{sec:training}).
Its robot-specific part carries the mechanism-level commands that have no human counterpart, arm-joint actions $\bm{a}^{\mathrm{ar}}_t \in \mathbb{R}^{14}$ and hand-joint commands $\bm{a}^{\mathrm{hd}}_t \in \mathbb{R}^{32}$, giving $\bm{a}_t = (\bm{a}^{\mathrm{wr}}_t, \bm{a}^{\mathrm{ar}}_t, \bm{a}^{\mathrm{hd}}_t, \bm{a}^{\mathrm{ft}}_t) \in \mathbb{R}^{88}$.
These sizes are those of our platform (App.~\ref{app:setup:layout}); \Ours relies only on the split between shared and robot-specific blocks.

\paragraph{Sources.}
Robot demonstrations are collected by bimanual teleoperation of a two-arm, two-hand platform, from which wrist poses, fingertip positions, and joint readings are recovered via forward kinematics.
Human demonstrations are drawn from EgoDex~\citep{hoque:iclr26}, an egocentric video dataset with tracked wrist and fingertip transforms; we transform trajectories into a body-centered frame to remove ego-motion, filter by hand visibility and velocity, and apply temporal smoothing before the same wrist-local retargeting (App.~\ref{app:data:preproc}). 
We then randomly mirror left and right hands to remove handedness bias.
Each episode is sliced into frame-level samples, each pairing the current observation and a language instruction with a future target of $H = 30$ steps.

\begin{figure}[t!]
    \begin{center}
    \includegraphics[width=1.00\textwidth]{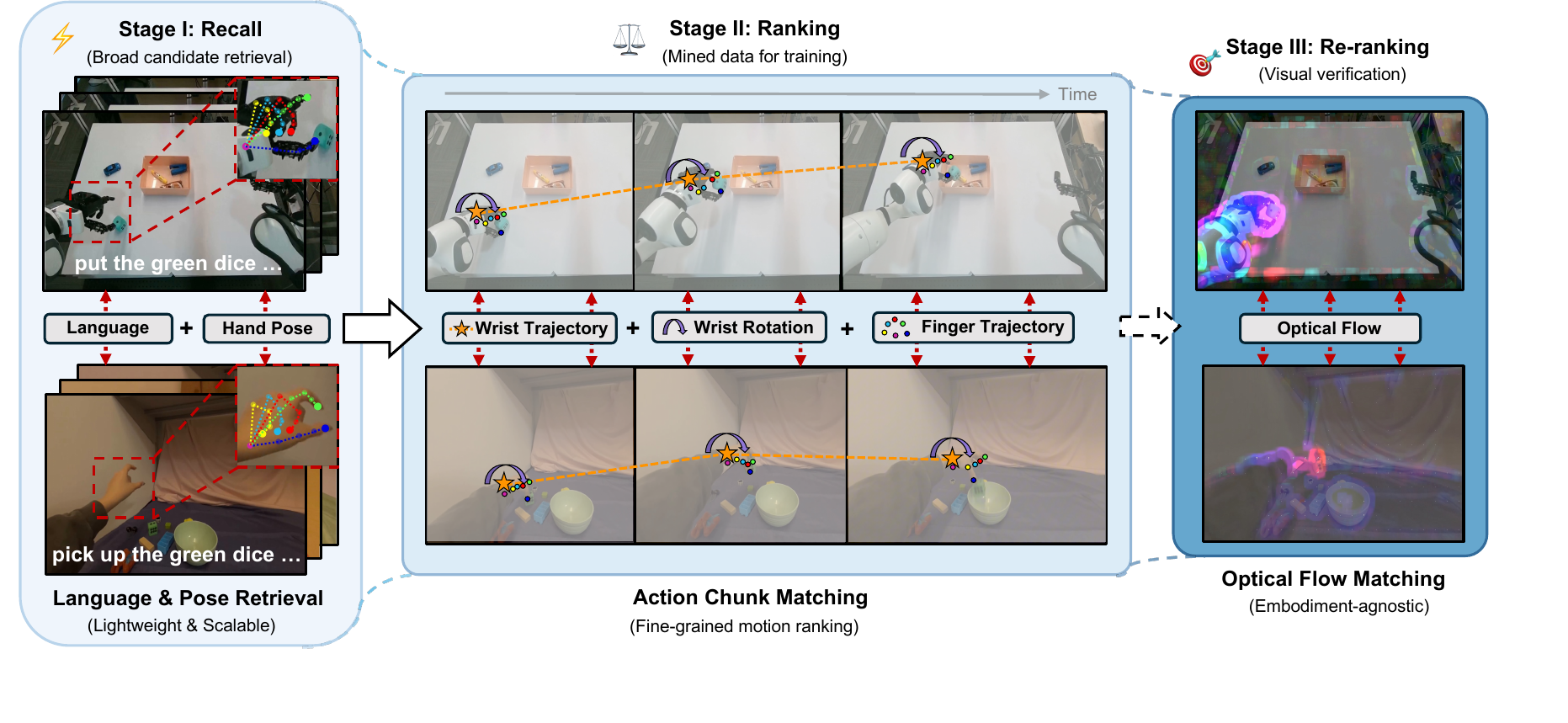}
    \end{center}
    \vspace{-6mm}
    \caption{\textbf{Overview of the \Ours pipeline.} Robot demonstrations (anchors, top) are matched against human candidates (bottom) through a cascade of increasing cost, denoted by deepening shades. The cascade \textbf{recalls} by language and hand pose and \textbf{ranks} by action-chunk similarity to select training data, then \textbf{re-ranks} by optical flow to verify it. Orange path: wrist position; purple arc: wrist rotation; colored dots: fingertip trajectories; red arrows: anchor--candidate comparison.}
\label{fig:pipeline}
\end{figure}

\subsection{Similarity-Based Data Mining}
\label{sec:mining}

Let $\mathcal{D}_r=\{\tau_n\}_{n=1}^{N}$ be the robot samples (anchors) and $\mathcal{D}_h=\{\tau_m\}_{m=1}^{M}$ the human pool (targets), with $M\gg N$.
Each sample $\tau$ pairs a language instruction with the observation and future action chunk of Sec.~\ref{sec:representation}, and retrieval runs per anchor.
Our goal is to retrieve the subset $\mathcal{D}_h^{*} \subset \mathcal{D}_h$ whose members are most similar to the anchors in hand configuration, language semantics, and motion dynamics.
Since exhaustive comparison over a pool of this size is infeasible, we adopt a cascade of increasing precision and cost (Fig.~\ref{fig:pipeline}; App.~\ref{app:method}).
Stage~I scans the full pool with lightweight signals and Stage~II re-scores only its candidates, forming a two-stage selector that produces the mined subset used for training.
Stage~III operates on pixels and provides embodiment-agnostic verification.

\paragraph{Stage I: Recall.} The recall stage retrieves a broad candidate set from two lightweight signals. For \emph{pose}, we $\ell_2$-normalize the initial fingertip configuration, removing hand scale, into a descriptor $\hat{\bm{p}} \in \mathbb{R}^{15}$ and retrieve by Euclidean distance, capturing similar initial grasp posture. For \emph{language}, we encode each instruction into an embedding $\bm{e} \in \mathbb{R}^{384}$ with an off-the-shelf sentence encoder and retrieve by cosine similarity, capturing task semantics. The two rankings are fused by reciprocal-rank fusion into a candidate set per anchor. Both signals are cheap enough to index the full pool.

\paragraph{Stage II: Ranking.} 
The ranking stage refines candidates by fine-grained motion similarity over the future action chunk, along four components derived from two trajectories: 
(i) the wrist trajectory $\bm{F}_{\mathrm{wr}} \in \mathbb{R}^{31\times 3}$ of one hand, the current position followed by the $30$ future ones, which preserves spatial structure;
(ii) a translation waveform $\bm{w}_{\mathrm{tr}} \in \mathbb{R}^{10}$ of the same hand, the per-step translation speed resampled from 30 steps to 10;
(iii) a rotation waveform $\bm{w}_{\mathrm{rot}} \in \mathbb{R}^{10}$, the per-step rotation speed obtained the same way from the wrist rotations;
and (iv) the fingertip trajectory $\bm{F}_{\mathrm{ft}} \in \mathbb{R}^{30\times 15}$, the $30$ future fingertip targets of one hand.
Each component is scored per hand and summed over the two hands, then sorted, giving four per-component ranks $r_{\mathrm{wr}}, r_{\mathrm{tr}}, r_{\mathrm{rot}}, r_{\mathrm{ft}}$; the fused score sums them with equal weight, $\rho = r_{\mathrm{wr}} + r_{\mathrm{tr}} + r_{\mathrm{rot}} + r_{\mathrm{ft}}$.
We then deduplicate by source trajectory, keeping only the highest-ranked entry
per source episode.
This prevents any single episode from dominating the mined subset $\mathcal{D}_h^*$, and the deduplicated output is what we use for VLA training.

\paragraph{Stage III: Re-ranking.}
The re-ranking stage provides an embodiment-agnostic verification using optical flow.
Unlike the preceding stages, which operate in the shared action space and therefore inherit any retargeting error, optical flow captures pixel-level motion patterns independent of hand morphology.
For each anchor--target pair, we aggregate dense flow into a clip-level descriptor that summarises motion direction and energy, then re-score by descriptor similarity, yielding a signal that does not depend on retargeting accuracy.
Because dense flow requires decoding and comparing video rather than cached action features, it is far costlier per pair than the action-space comparisons above, and we use it to analyse retrieval quality (Sec.~\ref{sec:visualization}) rather than to filter the training set.

\begin{figure}[t!]
    \begin{center}
    \includegraphics[width=1.00\textwidth]{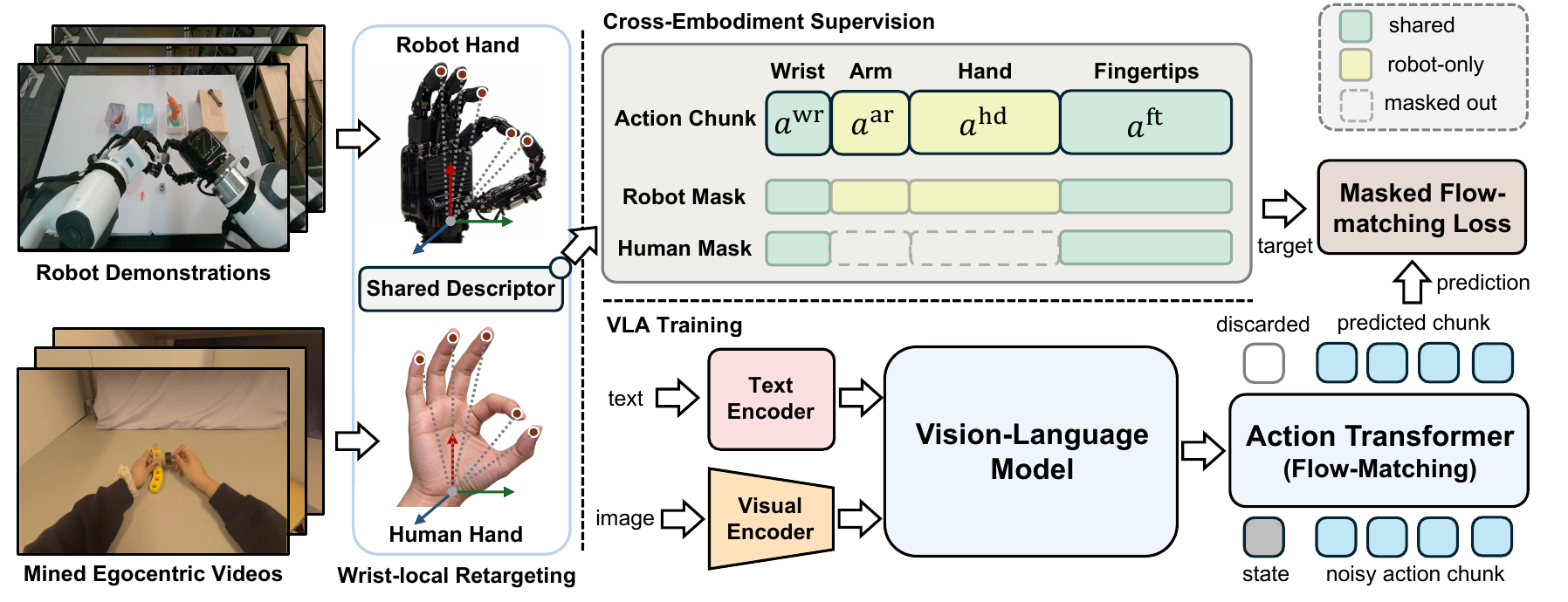}
    \end{center}
    \caption{\textbf{Cross-embodiment training with mined human data.} Robot demonstrations (top left) and mined egocentric videos (bottom left) are retargeted onto a shared descriptor and laid out in one unified action chunk. Green marks the embodiment-shared segments, yellow the robot-only ones, and dashed cells are excluded by the mask. The \textbf{upper lane} carries the supervision target built from the retargeted descriptor. The \textbf{lower lane} carries the model prediction, and the two meet at the masked flow-matching loss, which is computed only on the unmasked entries of the chunk.}
    \label{fig:training}
\end{figure}

\subsection{Cross-Embodiment Training}
\label{sec:training}

Mined human data enters training as an additional data source alongside the robot data, with no other change to the model or training procedure, so that data selection remains the only variable (Fig.~\ref{fig:training}).

\paragraph{Data streams.}
The mined subset $\mathcal{D}^{*}_h$ and the robot demonstrations $\mathcal{D}_r$ are loaded as two independently weighted data streams and sampled at a $1{:}1$ ratio (App.~\ref{app:setup:hparams}), rather than concatenated into one dataset.
Both are retargeted into the shared descriptor of Sec.~\ref{sec:representation}, so a single policy consumes them through the same inputs and outputs, with no source-specific module or embodiment flag.

\paragraph{Cross-embodiment supervision.}
All cross-embodiment difference is confined to the supervision side, where it takes the form of a mask.
Robot samples are supervised on all four segments of the action chunk, $88$ dimensions.
Human samples are supervised only on the embodiment-shared segments $\bm{a}^{\mathrm{wr}}$ and $\bm{a}^{\mathrm{ft}}$; the robot-specific segments $\bm{a}^{\mathrm{ar}}$ and $\bm{a}^{\mathrm{hd}}$ are filled with zero placeholders and excluded by a mask $\bm{m} \in \{0,1\}^{H\times 88}$. The same mask also removes future steps beyond the end of an episode, for both sources.
The mask is applied per sample rather than per batch, so a single batch can mix both sources freely.
Because the difference lives in the mask rather than in the network, human and robot data train a single model without any architectural change.

\paragraph{Architecture.}
The policy below the mask is an off-the-shelf $\pi_0$-like flow-matching VLA: a vision-language backbone encodes a single visual observation $\bm{I}_t$ and the language instruction $l$, a state adapter projects the proprioceptive state $\bm{s}_t \in \mathbb{R}^{86}$ into the backbone hidden space, and an action transformer denoises a noisy action chunk into the predicted chunk $\bm{a}_{t:t+H} \in \mathbb{R}^{H\times 88}$ over horizon $H$, laid out as in Sec.~\ref{sec:representation}.
The state pairs the robot's joint readings with the current fingertip configurations $\bm{p}_t$ of both hands; for human samples the joint slots are empty and zero-filled.
This path is identical for robot and human samples: the same encoders, the same backbone, and the same input and output dimensions.
\Ours leaves the backbone unchanged; App.~\ref{app:setup:model} details ours.

\paragraph{Training objective.} 
Supervision and prediction meet at the loss.
We adopt a flow-matching objective. Given a ground-truth action chunk $\bm{a} \in \mathbb{R}^{H\times 88}$, we sample noise $\bm{\epsilon}$ with entries drawn i.i.d.\ from $\mathcal{N}(0,1)$ and a timestep $\sigma \in [0,1]$, form an interpolated state $\bm{x}_\sigma \in \mathbb{R}^{H\times 88}$ with flow target $\bm{u}_\sigma \in \mathbb{R}^{H\times 88}$, and regress the predicted velocity $\hat{\bm{u}}_\sigma = f_\theta(\bm{x}_\sigma, \bm{I}_t, l, \bm{s}_t)$. The training loss is a masked mean squared error:
\begin{equation}
\mathcal{L} = \frac{\sum_{h,d} m_{h,d}\left([\hat{\bm{u}}_\sigma]_{h,d} - [\bm{u}_\sigma]_{h,d}\right)^2}{\sum_{h,d} m_{h,d}}
\label{eq:loss}
\end{equation}
where $h \in \{1,\dots,H\}$ indexes the prediction horizon, $d \in \{1,\dots,88\}$ indexes the action dimension, $[\hat{\bm{u}}_\sigma]_{h,d}$ and $[\bm{u}_\sigma]_{h,d}$ are the predicted and
target velocities at entry $(h,d)$, and $m_{h,d} \in \{0,1\}$ is the corresponding mask entry.
The mask $\bm{m}$ ensures human samples contribute gradients only through the $42$ wrist and fingertip dimensions, enabling training without architectural modification.

\section{Experiments}
\label{sec:experiment}
We design experiments to answer the following research questions:

\textbf{RQ1:} Does \Ours outperform a baseline trained on an equal amount of random human data?

\textbf{RQ2:} How does the benefit of \Ours scale with the amount of available robot demonstrations?

\textbf{RQ3:} How much does each stage of the \Ours cascade add to policy performance?

\textbf{RQ4:} What changes in the retrieved samples at each stage of the \Ours
cascade?

\subsection{Evaluation Tasks}
\label{sec:tasks}

\begin{figure*}[t]
\centering
\includegraphics[width=\textwidth]{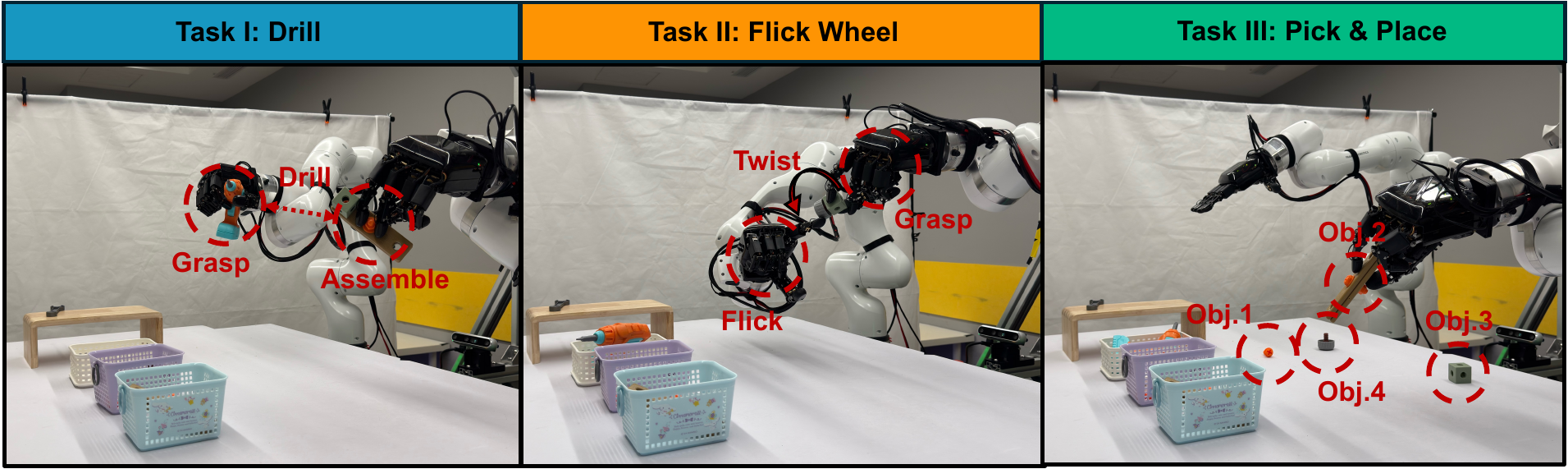}
\caption{\textbf{Overview of \Ours evaluation tasks.} We evaluate on three dexterous manipulation tasks: \textbf{Task~I: Drill} (grasp, assemble,  drill), \textbf{Task~II: Flick Wheel} (grasp, twist, flick), and \textbf{Task~III: Pick~\&~Place} (grasp and place 4 objects of diverse shapes). Red marks show key stages and objects.}
\label{fig:tasks}
\end{figure*}

As shown in Fig.~\ref{fig:tasks}, we design three dexterous manipulation tasks covering bimanual tool use, fine-grained finger dexterity, and precise grasping of multiple objects with diverse shapes.

\textbf{Task I: Drill} (max score $3$). The robot sequentially grasps the drill, aligns and assembles it with a fixture, and presses the trigger---evaluating tool use and multi-step coordination.

\textbf{Task II: Flick Wheel} (max score $3$). The robot grasps the assembly, twists the wheel with two fingers, and flicks it off with one finger. This is the most demanding in fine-grained finger dexterity, as the two-finger twist and one-finger flick stages require precise fingertip coordination.

\textbf{Task III: Pick \& Place} (max score $4$). Four objects of diverse geometries are scattered on the table; the robot picks and places each at a designated location, testing precise grasps across object shapes.

Stages are sequential except in Pick \& Place, whose objects are scored independently.
Sub-tasks earn partial credit per completed step, and success rate is total score over the maximum; each task is evaluated over $10$ trials ($2$ rounds $\times$ $5$ layouts) by an independent team blind to the method (App.~\ref{app:eval}).

\subsection{Implementation Details}
\label{sec:implementation}

\begin{table}[t]
\centering
\caption{%
\textbf{Real-world evaluation on dexterous manipulation tasks (RQ1).}
Sub-task scores (\mbox{$0$--$1$}) are means over $10$ trials ($2$ rounds $\times$ $5$); $\pm$ is the standard deviation over the two round-level means ($n=2$; App.~\ref{app:eval:aggregation}).
Total $=$ sum of sub-task scores; Success Rate (SR) $=$ Total\,/\,max\,$\times100\%$, i.e., a normalized progress score.
Best in \textbf{bold}.
\mbox{$\Delta=$ SiMDex $-$ GR-Dexter}~\citep{wen:arxiv25}, the identical model trained with an equal amount of randomly sampled human data.
GR-Dexter*, not used in $\Delta$, is trained on a fixed random subset of the same size as the mined set (${\sim}1.49$M samples), drawn once from the full pool.
Totals, rates, and $\Delta$ are computed from unrounded scores, so $\Delta$ may differ by at most one unit in the last digit from the difference of the rounded entries.}
\label{tab:main}
{\small
\setlength{\tabcolsep}{4pt}
\renewcommand{\arraystretch}{1.1}
\newlength{\tabsubw}\settowidth{\tabsubw}{\textbf{0.00}\tiny$\pm$.00}
\newlength{\tabsubhalf}\setlength{\tabsubhalf}{\dimexpr 0.5\tabsubw+\tabcolsep+\cmidrulekern\relax}
\newcommand{\threesub}[3]{\multicolumn{4}{c}{\makebox[\tabsubw]{#1}\hspace{2\tabcolsep}\makebox[\tabsubw]{#2}\hspace{2\tabcolsep}\makebox[\tabsubw]{#3}}}
\begin{tabular}{llcccccc}
\toprule
Task & Method & \multicolumn{4}{c}{Sub-task Scores} & Total & Success Rate(\%)$\uparrow$ \\
\midrule
\multirow{5}{*}{\shortstack[l]{Drill\\\scriptsize(max 3)}}
 & & \threesub{Grasp}{Assem.}{Drill} & & \\
 \cmidrule(l{\tabsubhalf}r{\tabsubhalf}){3-6}
 & GR-Dexter* & \threesub{0.43\tiny$\pm$.14}{0.40\tiny$\pm$.00}{0.10\tiny$\pm$.05} & 0.93\tiny$\pm$.09 & 31.1 \\
 & GR-Dexter & \threesub{\textbf{0.80}\tiny$\pm$.00}{\textbf{0.80}\tiny$\pm$.00}{\textbf{0.33}\tiny$\pm$.00} & \textbf{1.93}\tiny$\pm$.00 & \textbf{64.5} \\
 & SiMDex (Ours) & \threesub{0.73\tiny$\pm$.10}{0.63\tiny$\pm$.33}{0.27\tiny$\pm$.28} & 1.63\tiny$\pm$.71 & 54.5 \\
 & $\Delta$ & \threesub{\dneg{$-$0.07}}{\dneg{$-$0.17}}{\dneg{$-$0.07}} & \dneg{$-$0.30} & \dneg{$-$10.0} \\
\midrule
\multirow{5}{*}{\shortstack[l]{Flick Wheel\\\scriptsize(max 3)}}
 & & \threesub{Grasp}{Twist}{Flick} & & \\
 \cmidrule(l{\tabsubhalf}r{\tabsubhalf}){3-6}
 & GR-Dexter* & \threesub{0.65\tiny$\pm$.21}{0.27\tiny$\pm$.28}{0.05\tiny$\pm$.07} & 0.97\tiny$\pm$.57 & 32.2 \\
 & GR-Dexter & \threesub{0.60\tiny$\pm$.14}{0.13\tiny$\pm$.19}{0.00\tiny$\pm$.00} & 0.73\tiny$\pm$.33 & 24.5 \\
 & SiMDex (Ours) & \threesub{\textbf{0.85}\tiny$\pm$.07}{\textbf{0.47}\tiny$\pm$.09}{\textbf{0.10}\tiny$\pm$.00} & \textbf{1.42}\tiny$\pm$.17 & \textbf{47.2} \\
 & $\Delta$ & \threesub{\dpos{$+$0.25}}{\dpos{$+$0.33}}{\dpos{$+$0.10}} & \dpos{$+$0.68} & \dpos{$+$22.8} \\
\midrule
\multirow{5}{*}{\shortstack[l]{Pick \& Place\\\scriptsize(max 4)}}
 & & \makebox[\tabsubw]{Obj.1} & \makebox[\tabsubw]{Obj.2} & \makebox[\tabsubw]{Obj.3} & \makebox[\tabsubw]{Obj.4} & & \\
 \cmidrule(lr){3-6}
 & GR-Dexter* & 0.26\tiny$\pm$.00 & 0.70\tiny$\pm$.14 & 0.50\tiny$\pm$.14 & 0.57\tiny$\pm$.05 & 2.03\tiny$\pm$.24 & 50.7 \\
 & GR-Dexter & 0.30\tiny$\pm$.05 & 0.67\tiny$\pm$.09 & 0.53\tiny$\pm$.09 & 0.67\tiny$\pm$.09 & 2.16\tiny$\pm$.05 & 54.0 \\
 & SiMDex (Ours) & \textbf{0.67}\tiny$\pm$.19 & \textbf{0.87}\tiny$\pm$.00 & \textbf{0.97}\tiny$\pm$.05 & \textbf{0.83}\tiny$\pm$.05 & \textbf{3.33}\tiny$\pm$.19 & \textbf{83.3} \\
 & $\Delta$ & \dpos{$+$0.37} & \dpos{$+$0.20} & \dpos{$+$0.44} & \dpos{$+$0.17} & \dpos{$+$1.17} & \dpos{$+$29.4} \\
\midrule
\multicolumn{6}{l}{\textbf{Overall Success Rate(\%)}} & \multicolumn{2}{c}{47.7 $\rightarrow$ \textbf{61.7}~~\dpos{$+$14.0}} \\
\bottomrule
\end{tabular}
}
\end{table}   %

\textbf{Base model.} Our base model follows GR-Dexter~\citep{wen:arxiv25}, a $\pi_0$-like flow-matching VLA for dexterous manipulation.
Its pre-training data include EgoDex, the egocentric corpus that \Ours re-mines during post-training.
\Ours modifies only the human-data source, leaving the architecture and training procedure unchanged;
the baseline (\Baseline) uses the identical model and an equal amount of randomly sampled human data, so only the selection differs.

\textbf{Data.} Robot demonstrations are collected by bimanual teleoperation of a two-arm, two-hand platform.
From the raw collection we curate a high-quality subset of $\sim$1.35M frame-level samples ($\sim$12.4 hours; App.~\ref{app:data:robot}), which serves as the default robot-data budget ($1\times$) throughout the paper; all results in Table~\ref{tab:main} use this setting.
The human pool is derived from EgoDex~\citep{hoque:iclr26}, an egocentric manipulation dataset of $829$ hours across $194$ tabletop tasks. After the preprocessing of Sec.~\ref{sec:representation}, it yields $32{,}034{,}551$ frame-level samples ($\sim$32M; App.~\ref{app:data:pool}) over $164{,}959$ episodes, sliced into sliding-window samples (each a $\sim$1\,s, $30$-step window).
From this pool, \Ours retrieves $\sim$1.49M samples (under $5\%$ of the pool). \Baseline draws human samples at random from the full pool, seeing as many as \Ours without being limited to a fixed subset (App.~\ref{app:setup:budget}).

\textbf{Training.} 
For the main results we train with a 1:1 mixture of robot and human samples for 40K steps, roughly two passes over each source.
All hyperparameters are identical between \Ours and the baseline, isolating data selection rather than model capacity or training budget (App.~\ref{app:setup:budget}).

\subsection{Main Results}
\label{sec:main_results}

Table~\ref{tab:main} reports real-world results under our standard setting.
Mining task-relevant human data lifts the overall success rate from $47.7\%$ to $61.7\%$ ($+14.0$) with only $\sim$1.49M retrieved samples (under $5\%$ of the pool), showing that targeted retrieval outperforms an equal amount of randomly sampled human data.
This gain is statistically significant: over 30 pairs of trials that share task, layout and round, a two-sided paired permutation test gives $p = 0.022$ (95\% bootstrap CI $[+2.7, +24.9]$; App.~\ref{app:eval:aggregation}).
The gain is larger against GR-Dexter*, which is trained on a random subset of the same size as the mined set (Tables~\ref{tab:main} and~\ref{tab:equal-size}).
For context, \Ours also scores above policies post-trained on robot data alone, matched in training steps or in robot-data passes (App.~\ref{app:robot-only}).

\textbf{Per task.} \Ours improves two of three complex tasks by large margins.
On \emph{Flick Wheel}, it nearly doubles the success rate ($24.5\rightarrow47.2$, $+22.8$). The gain comes from completing more stages, and only \Ours reaches the final Flick stage (App.~\ref{sec:supp-cond-pass}).
On \emph{Pick \& Place}, it improves $54.0\rightarrow83.3$ ($+29.4$) with consistent gains on all four objects.
On \emph{Drill}, \Ours is lower than the baseline (54.5 vs.\ 64.5).
The Drill gap arises at Grasp and Assembly, not at drilling (App.~\ref{sec:supp-cond-pass}), and is not significant (paired $p = 0.51$). When the four budgets are pooled, the two methods are level on Drill (App.~\ref{sec:supp-scaling-full}).

\subsection{Data Scaling Ablation}
\label{sec:scaling}

\begin{figure*}[t]
\centering
\includegraphics[width=\linewidth]{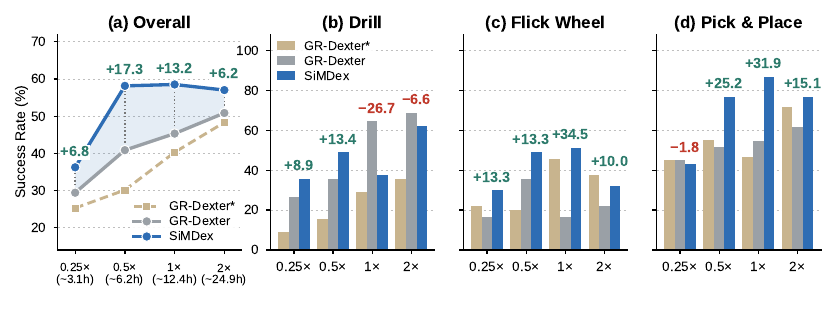}
\caption{%
\textbf{Data scaling ablation (RQ2).}
\textbf{(a)} Overall SR vs.\ robot-data scale: \Ours stays at $\sim$57--59\% from $0.5\times$ to $2\times$ while the baseline drops with less data (largest gain $+17.3$ at $0.5\times$).
\textbf{(b--d)} Per-task SR.
All panels include GR-Dexter*, and $\Delta=$ \Ours $-$ GR-Dexter.
All results use one round of five trials per task (at $1\times$, the first round of Table~\ref{tab:main}). Per-sub-task numbers are in App.~\ref{sec:supp-scaling-full}.}
\label{fig:scaling}
\end{figure*}

Since robot data is costly, we vary the robot-data budget from $0.25\times$ ($\sim$3.1h) to $2\times$ ($\sim$24.9h) while keeping the mined human data fixed (Fig.~\ref{fig:scaling}).
This subset is mined once with the $1\times$ anchors and reused at every budget, and the $2\times$ budget adds lower-quality early teleoperation data (App.~\ref{app:scaling}).

\textbf{Gains across budgets.} \Ours outperforms the baseline at every scale (Fig.~\ref{fig:scaling}a) and stays at $\sim$57--59\% from $0.5\times$ to $2\times$, while the baseline falls from $50.9\%$ to $40.8\%$.
The largest gain ($+17.3$) appears at $0.5\times$, and at $0.25\times$ \Ours still improves by $+6.8$.
Against GR-Dexter*, \Ours is higher at every budget, by $+17.8$ when the four budgets are pooled ($p<0.0001$, 95\% CI $[+11.8, +24.1]$), with the full table and test details in App.~\ref{sec:supp-scaling-full}.
Since the subset is reused, its anchors include robot demonstrations absent from the smaller budgets. Re-mining per budget is left to future work.

\textbf{Per task.} The task panels (Fig.~\ref{fig:scaling}b--d) show that \Ours gains on Flick Wheel at every budget and on Pick \& Place at every budget except $0.25\times$, while the sign of the Drill gap varies.
With a single round of five trials per point, we read per-task differences at a single budget as indicative only.

\subsection{\texorpdfstring{What Each Stage Contributes}{What Each Stage Contributes}}
\label{sec:stage-ablation}

\begin{table}[t]
\centering
\caption{%
\textbf{What each stage contributes, at $1\times$ (RQ3).}
In the middle three rows, the model of Table~\ref{tab:main} is post-trained on the cascade output up to the given stage, keeping the top $50$ per anchor, with the same training steps and 1:1 mixture.
The first and last rows repeat Table~\ref{tab:main}.
SR in \%.
}
\label{tab:stage-ablation}
{\small
\setlength{\tabcolsep}{4pt}
\begin{tabular}{lccccc}
\toprule
Method & \makecell{Distinct\\samples} & Drill & \makecell{Flick\\Wheel} & \makecell{Pick \&\\Place} & Overall \\
\midrule
GR-Dexter & drawn from $\sim$32M & 64.5 & 24.5 & 54.0 & 47.7 \\
\midrule
Recall (Stage I) only & 0.14M & 2.2 & 60.5 & 59.1 & 40.6 \\
+ Ranking (Stage II) & 1.05M & 31.1 & 58.9 & 85.0 & 58.3 \\
+ Re-ranking (Stage III) as a filter & 1.03M & 64.5 & 37.8 & 83.3 & 61.9 \\
\midrule
SiMDex (Ours) & 1.49M & 54.5 & 47.2 & 83.3 & 61.7 \\
\bottomrule
\end{tabular}}
\end{table}

For RQ3 (Table~\ref{tab:stage-ablation}), we post-train the same model at $1\times$ on the cascade output up to each stage.
Each run keeps the top $50$ samples per anchor, so that Re-ranking, which only reorders its input, changes which samples are kept.
Mining uses fewer anchors than Table~\ref{tab:main} (App.~\ref{app:stagewise}). Three findings emerge.

\textbf{Language and initial pose are not enough.}
Recall alone is no better than \Baseline (40.6 vs.\ 47.7, $p = 0.44$).
Despite the same quota, it also keeps far fewer distinct samples (0.14M vs.\ 1.05M after Ranking), since anchors that share instructions and initial poses tend to recall the same clips.

\textbf{Hand-motion ranking drives the gain.}
Adding Ranking, which refines the recalled candidates by fine-grained motion similarity, lifts the overall SR from 40.6 to 58.3 ($+17.7$, $p = 0.017$).

\textbf{Optical flow serves verification, not selection.}
Used as a filter on the Ranking output, Re-ranking does not change the overall SR significantly (61.9 vs.\ 58.3, $p = 0.73$).
\Ours therefore selects training data with Recall and Ranking alone and verifies retrieval with optical flow (Sec.~\ref{sec:visualization}).

\subsection{Visualization}
\label{sec:visualization}

\begin{figure}[t!]
    \begin{center}
    \includegraphics[width=0.9\textwidth]{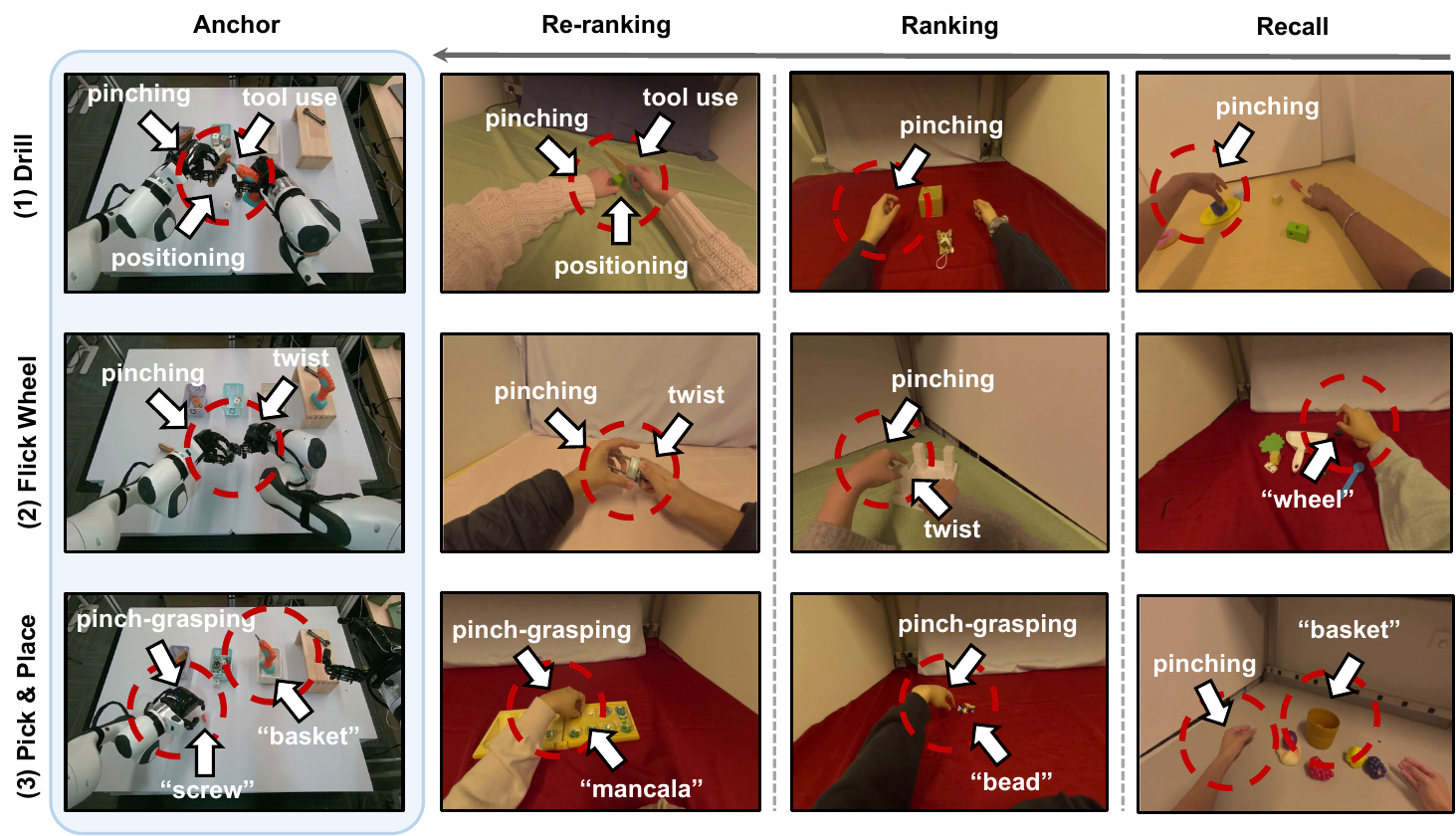}
    \end{center}
      \caption{%
\textbf{Qualitative visualization of the \Ours retrieval cascade (RQ4).}
Each row shows a robot anchor (left) and the top retrieved human sample after each stage, from Recall (right) to Re-ranking.
Plain labels name hand actions, quoted labels name objects, and red circles mark interaction regions.
Re-ranking then adds a quality check on pixels via optical flow, flagging samples that match the anchor in the action space but move differently in the egocentric video.}
    \label{fig:visualization}
\end{figure}

For RQ4, Fig.~\ref{fig:visualization} shows, for three anchors, the top retrieved sample after each stage; in these examples, the match to the anchor tightens step by step from Recall through Ranking to Re-ranking.

\textbf{Recall} matches coarse cues only: it retrieves samples with a similar initial pose or language-relevant objects (a ``wheel'', a ``basket''), but the actions are often unrelated to the target---simple planar grasping rather than tool use or twisting.
\textbf{Ranking} aligns fine-grained motion: pinching and twisting patterns, finger curvature, and rotation direction come closer to the anchor. Yet samples with distant hand positions can still rank high when their kinematic values happen to match.
\textbf{Re-ranking} reorders these candidates by optical flow; its top samples, already in the training subset, match the anchor's motion: a left hand stabilizing while the right aligns a tool (Drill), a coordinated stabilize-and-twist (Flick Wheel), and sequential precision grasping of small items (Pick \& Place).
In short, recall brings in clips with related objects or initial poses, ranking aligns their motion with the anchor, and re-ranking checks that motion on pixels. More examples, including failure cases, are in App.~\ref{app:mined}.

\section{Conclusion}
\label{sec:conclusion}
We presented \Ours, a similarity-based data mining framework that retrieves task-relevant human demonstrations from large-scale egocentric video to improve VLA training for dexterous manipulation.
Cast as a retrieval problem, \Ours uses recall and ranking to mine a small, high-quality subset from a pool of tens of millions of human samples and re-ranking to verify it, modifying only the data source while leaving the VLA architecture and training unchanged.
Across three real-world tasks, mining lifts the overall success rate from $47.7\%$ to $61.7\%$ using under $5\%$ of the pool and stays above both random baselines at every robot-data budget from $0.25\times$ to $2\times$. An ablation traces the gain to ranking by hand motion.
This points to selective human-data mining as a practical complement to the costly collection of more teleoperated robot demonstrations.

\section{Limitations}
\label{sec:limitations}

\textbf{Scope.}  Our robot data covers one scenario (industrial assembly), three
tasks, and $\sim$12.4 hours of demonstrations, so our claims concern these tasks on the reported platform, at the amounts of human data reported.
Whether mining for target tasks affects unrelated tasks remains an open question.

\textbf{What the comparison isolates.} Our controls hold the human-data budget fixed and vary only how that data is selected, since the value of human data at scale is documented in prior work~\citep{nvidia:arxiv26}.
Our claims therefore concern selection---targeted retrieval beats both random baselines---not the value of human data in general. Robot-only references are reported for context in App.~\ref{sec:supp-scaling-full}.

\textbf{Dependence on pool coverage.} \Ours depends on pool coverage: no EgoDex task involves drilling, the drilling stage of Drill has no counterpart among the retrieved clips (App.~\ref{app:mined-failures}), and mining brings no gain there.
Averaged over the budgets, only Drill gains from a larger random pool (GR-Dexter vs.\ GR-Dexter*), so both the amount and coverage of human data bear on it (App.~\ref{sec:supp-scaling-full}).

\textbf{Kinematic-only similarity.} Our similarity is purely kinematic (wrist 6D pose and fingertip positions): it captures motion but not contact forces, object state, or interaction semantics. The ranking stage can therefore mismatch---e.g., pairing the robot's nearly static drilling with oscillating tool-use clips---an error that re-ranking can flag on pixels via optical flow (App.~\ref{app:mined-failures}).

\textbf{Scope of the ablation.} Our stage-wise ablation tests Ranking (Stage II) as a whole. Single ranking components and language-only or task-label filters are left for future work.
\section{Future Work}
\label{sec:future_work}

These limitations point to a natural next step: moving from one-shot, offline retrieval toward closed-loop, adaptive mining.
In the current pipeline, \emph{what} to retrieve is fixed before training, yet our scaling analysis suggests that the most useful human data may depend on the robot-data budget and on which skills the policy has yet to master.
The policy's failure modes could steer subsequent retrieval, and the similarity signal could capture contact and object state.
Closing this loop would let egocentric collection be exploited not once, but continually throughout policy learning.
Ultimately we see \Ours as a step toward a recommender for robot skills, where learning a new skill is not a fresh data-collection burden but a query against experience that already exists.

\newpage
\bibliographystyle{iclr2027_conference}
\bibliography{iclr2027_conference,refs/ref_base,refs/ref_hands,refs/ref_data,refs/ref_VLA,refs/ref_human_data,refs/ref_data_curation}

\newpage
\appendix
\section*{Appendix}

\begin{center}
  \includegraphics[width=0.95\linewidth]{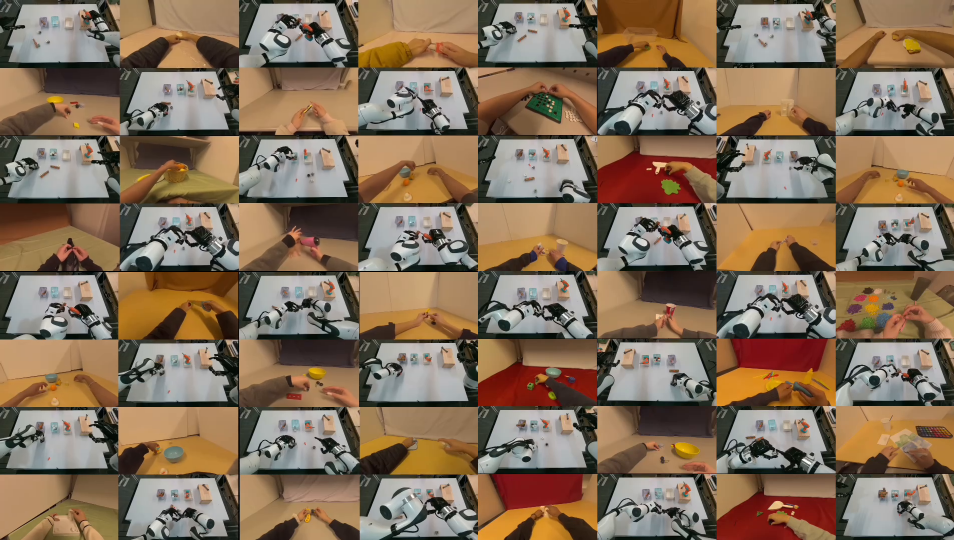}
  \vspace{-0.5em}
  \captionof{figure}{\textbf{Robot anchors and the human samples mined by \Ours.}
  Each robot tile is paired with the adjacent human tile mined for it from EgoDex~\citep{hoque:iclr26}.}
  \label{fig:appendix_hero}
\end{center}

Each pair in Figure~\ref{fig:appendix_hero} reflects three ideas behind \Ours:
\begin{itemize}[leftmargin=*, itemsep=2pt, topsep=3pt]
  \item \textbf{Relevance lies in motion, not labels.}
  The two tiles of a pair differ in embodiment, scene, and activity, yet share the motion itself: a pinch, a wrist twist, or a hand-over.
  \item \textbf{Labels can mislead.}
  Grasps mined for a sub-task come from recordings of other activities, so a pool that looks unrelated by its labels can still hold the motion the robot needs (App.~\ref{app:mined}).
  \item \textbf{Retrieval must fit a budget.}
  Like a recommender, \Ours{} surfaces useful samples from tens of millions under a strict compute budget, so it uses a cascade of cheap signals and keeps the costly pixel-level signal (optical flow) for verification only (Stage~III, Sec.~\ref{sec:mining}).
\end{itemize}
 
The rest of the appendix covers method details (App.~\ref{app:method}), data and the human pool (App.~\ref{app:data}), what gets mined (App.~\ref{app:mined}), experimental setup (App.~\ref{app:setup}), evaluation protocol (App.~\ref{app:eval}), additional results (App.~\ref{app:results}), and the details of the stage-wise ablation of Sec.~\ref{sec:stage-ablation} (App.~\ref{app:stagewise}).
\newpage

\section{Method Details}
\label{app:method}

This appendix expands the method of Sec.~\ref{sec:method} with the full set of definitions, hyperparameters, and per-stage formulas that were abbreviated in the main paper.
We keep the notation of the main text throughout, introducing a symbol only where the appendix needs to name something that the main text refers to in words.
Nothing below modifies the method of Sec.~\ref{sec:method}: every added quantity is a concrete realization of a component already described there.

\subsection{Unified Morphology-Agnostic Representation}
\label{app:representation}

\paragraph{From single-hand to bimanual action.}
The main text defines, for a single hand, the wrist-local fingertip configuration $\bm{p}_t \in \mathbb{R}^{15}$, the wrist action $\bm{d}_t \in \mathbb{R}^{6}$ for the $k$-th future step, and the fingertip target $\bm{f}_t := \bm{p}_{t+k} \in \mathbb{R}^{15}$ predicted at that same future step.
We make the single-hand~$\rightarrow$~bimanual composition explicit here and give an explicit symbol to the 42 embodiment-shared dimensions, which the main text refers to only by their count.
Writing the left and right hands with superscripts L and R, the two per-hand quantities are

\begin{equation}
\bm{d}^{\mathrm{L}}_t,\ \bm{d}^{\mathrm{R}}_t \in \mathbb{R}^{6},
\qquad
\bm{f}^{\mathrm{L}}_t,\ \bm{f}^{\mathrm{R}}_t \in \mathbb{R}^{15},
\end{equation}

and the two embodiment-shared segments of the main text, $\bm{a}^{\mathrm{wr}}_t \in \mathbb{R}^{12}$ and $\bm{a}^{\mathrm{ft}}_t \in \mathbb{R}^{30}$, are exactly their per-hand concatenations.
We write the two segments together, in the order in which they appear in $\bm{a}_t$, as the shared action

\begin{equation}
\bm{a}^{\mathrm{sh}}_t = \bigl(\bm{a}^{\mathrm{wr}}_t,\ \bm{a}^{\mathrm{ft}}_t\bigr)
\in \mathbb{R}^{42}.
\label{eq:shared-action}
\end{equation}

Its 42 dimensions decompose as $2 \times 6 = 12$ for the two wrists and $2 \times 15 = 30$ for the ten fingertips; they are the 42 wrist and fingertip dimensions through which human samples contribute gradients in Sec.~\ref{sec:training}.
The symbol $\bm{a}^{\mathrm{sh}}_t$ is a notational convenience only: it does not change the action layout of the main text, but names those components of $\bm{a}_t$ that are embodiment-shared, because the action-space comparison of mining operates on them alone.
The two hands are handled symmetrically and combined as described in App.~\ref{app:bimanual}.

\paragraph{Anchor and target sets.}
Following the main text, let $\mathcal{D}_r = \{\tau_n\}_{n=1}^{N}$ be the robot samples (anchors) and $\mathcal{D}_h = \{\tau_m\}_{m=1}^{M}$ the human target pool, with $M \gg N$.
Each sample $\tau$ pairs a language instruction $l$ with an observation and a future action chunk; mining uses only $l$, the observed fingertip configuration $\bm{p}_t$, and the shared part $\bm{a}^{\mathrm{sh}}_{t:t+H} \in \mathbb{R}^{H \times 42}$ of that chunk.
Anchors and targets are mapped onto the identical representation of Sec.~\ref{sec:representation} and all features are extracted by the \emph{same} operators, so that similarities are computed in one common space; this symmetry is what makes cross-embodiment comparison well defined and is assumed throughout.
Throughout, we write $s(\bm{z}, \bm{z}')$ for the cosine similarity between two real vectors $\bm{z}$ and $\bm{z}'$ of the same length, so that $s(\bm{z}, \bm{z}') \in [-1, 1]$.

\subsection{Stage I: Recall}
\label{app:recall}

Recall combines two lightweight per-anchor rankings, one from initial hand pose and one from language, into a single candidate set.
Fix an anchor $\tau \in \mathcal{D}_r$; all ranks below are taken over targets $\tau' \in \mathcal{D}_h$ with respect to this anchor, and every anchor is processed independently.

\paragraph{Pose ranking.}
As in the main text, for a single hand we $\ell_2$-normalize the observed fingertip configuration $\bm{p}_t$, removing hand scale, to obtain the pose descriptor

\begin{equation}
\hat{\bm{p}} = \frac{\bm{p}_t}{\lVert \bm{p}_t \rVert_2} \in \mathbb{R}^{15}.
\end{equation}

Targets are retrieved by Euclidean distance, summed over the two hands,

\begin{equation}
\delta_{\mathrm{pose}}
= \bigl\lVert \hat{\bm{p}}^{\mathrm{L}} - \hat{\bm{p}}^{\prime\mathrm{L}} \bigr\rVert_2
+ \bigl\lVert \hat{\bm{p}}^{\mathrm{R}} - \hat{\bm{p}}^{\prime\mathrm{R}} \bigr\rVert_2 ,
\label{eq:pose-dist}
\end{equation}

where $\hat{\bm{p}}$ and $\hat{\bm{p}}'$ are the anchor and target pose descriptors.
For a single hand, both descriptors are unit vectors, so $\lVert \hat{\bm{p}} - \hat{\bm{p}}' \rVert_2^2 = 2 - 2\,s(\hat{\bm{p}}, \hat{\bm{p}}')$ and ordering by Euclidean distance is equivalent to ordering by cosine similarity; we use the Euclidean form to match the main text.
Ordering targets by increasing $\delta_{\mathrm{pose}}$ yields a pose rank $r_{\mathrm{pose}}(\tau') \in \{1, \dots, M\}$ for each target $\tau'$ (smaller is more similar).
The low-dimensional descriptor keeps this ranking cheap even over the full pool.

\paragraph{Language ranking.}
Each instruction $l$ is encoded into a sentence embedding $\bm{e} \in \mathbb{R}^{384}$ by a pretrained sentence encoder.\footnote{We use the public \texttt{all-MiniLM-L6-v2} encoder from Sentence-Transformers~\citep{reimers:emnlp19}, built on MiniLM~\citep{wang:neurips20}; it produces 384-dimensional embeddings and is not fine-tuned.}
Ordering targets by decreasing $s(\bm{e}, \bm{e}')$, where $\bm{e}'$ is the target embedding, yields a language rank $r_{\mathrm{lang}}(\tau') \in \{1, \dots, M\}$ (smaller is more similar).

\paragraph{Rank fusion.}
The two rankings are merged by reciprocal-rank fusion, giving the fused recall score

\begin{equation}
s_{\mathrm{rrf}}(\tau')
= \frac{1}{2}\,\frac{1}{\kappa + r_{\mathrm{pose}}(\tau')}
+ \frac{1}{2}\,\frac{1}{\kappa + r_{\mathrm{lang}}(\tau')},
\qquad \kappa = 60 ,
\label{eq:rrf}
\end{equation}

where $\kappa \in \mathbb{N}$ is the rank-fusion constant that controls how sharply the contribution of a ranking decays with rank: a larger $\kappa$ flattens the difference between top ranks, so that a target favored by \emph{both} the pose and language rankings is rewarded over one ranked first by a single signal alone.
We set $\kappa = 60$, the standard default for reciprocal-rank fusion.
We fuse ranks rather than scores because the two signals are not commensurable: pose is a Euclidean distance in $\mathbb{R}^{15}$ and language a cosine similarity in $\mathbb{R}^{384}$, so their scores cannot be added directly.
The reciprocal transform further bounds the contribution of each ranking by $\tfrac{1}{2(\kappa + 1)}$, so that neither signal can dominate the fused ordering through the long tail of its own rank distribution.

\paragraph{Candidate cutoff.}
The fused list is narrowed to the top $K_{\mathrm{I}} = 10{,}000$ targets.
The cutoff is set by compute: it keeps the costlier framewise matching of Stage~II, which compares each candidate over the whole chunk, affordable.
Writing $r_{\mathrm{recall}}(\tau')$ for the rank of $\tau'$ under decreasing $s_{\mathrm{rrf}}$, the retained candidate set is

\begin{equation}
\mathcal{C}_{\mathrm{I}}(\tau)
= \bigl\{\, \tau' \in \mathcal{D}_h \;\bigm|\; r_{\mathrm{recall}}(\tau') \le K_{\mathrm{I}} \,\bigr\},
\qquad \bigl|\mathcal{C}_{\mathrm{I}}(\tau)\bigr| = K_{\mathrm{I}} .
\end{equation}

\subsection{Stage II: Ranking}
\label{app:ranking}

Ranking refines $\mathcal{C}_{\mathrm{I}}(\tau)$ by fine-grained motion similarity along the four components named in the main text.
We give the concrete form of each component and of their fusion.

\paragraph{Motion components.}
The four components of the main text are two framewise trajectories, which preserve spatial structure, and two fixed-length waveforms, which summarize motion rhythm; all are computed per hand.
The wrist trajectory $\bm{F}_{\mathrm{wr}} \in \mathbb{R}^{31 \times 3}$ is the current wrist position followed by the 30 future ones; we index its rows from $0$, so that row $0$ is the current position, which is the origin in wrist-local coordinates.
The fingertip trajectory $\bm{F}_{\mathrm{ft}} \in \mathbb{R}^{30 \times 15}$ holds the fingertip targets of one hand over the $H = 30$ future steps, with the five fingertips stacked along the feature dimension ($5 \times 3 = 15$).
The translation waveform $\bm{w}_{\mathrm{tr}} \in \mathbb{R}^{10}$ is the per-step translation speed resampled from 30 steps to 10 samples.
Concretely, we prepend a zero frame to the future wrist translation sequence and take consecutive differences to obtain per-step increments.
The norm of each increment gives a scalar speed sequence of length 30.
We then normalize that sequence in magnitude, build a phase variable from its cumulative path length, and resample along that phase to the fixed length of 10.
The rotation waveform $\bm{w}_{\mathrm{rot}} \in \mathbb{R}^{10}$ is obtained by applying the same operator to the per-step wrist rotation increments.
Normalization and path-length resampling make a waveform describe how speed varies along the path,
not how fast or how long the motion was.

\paragraph{Per-component similarity.}
The two waveforms are scored by cosine similarity on the 10-dimensional vectors,

\begin{equation}
s_{\mathrm{tr}} = s(\bm{w}_{\mathrm{tr}}, \bm{w}'_{\mathrm{tr}}),
\qquad
s_{\mathrm{rot}} = s(\bm{w}_{\mathrm{rot}}, \bm{w}'_{\mathrm{rot}}).
\end{equation}

The two trajectories are scored by a framewise similarity averaged over the $H = 30$ future steps.
Writing $\bm{F}_{i:}$ for the row of a trajectory matrix $\bm{F}$ at future step $i$, the similarity between an anchor trajectory $\bm{F}$ and a target trajectory $\bm{F}'$ is

\begin{equation}
s_{\mathrm{traj}}(\bm{F}, \bm{F}') = \frac{1}{H}\sum_{i=1}^{H} s(\bm{F}_{i:},\ \bm{F}'_{i:}),
\end{equation}

applied to the wrist and the fingertip trajectory to give $s_{\mathrm{wr}}$ and $s_{\mathrm{ft}}$.
Row $0$ of the wrist trajectory is the origin for every sample; it carries no information and its cosine similarity is undefined, so it is left out of the average.
Including it under any fixed convention would rescale and shift every score in the same way and leave the ranking unchanged.
As stated in the main text, every component is scored per hand and then summed over the two hands to give a single bimanual component score.

\paragraph{Rank-sum fusion and deduplication.}
Each of the four components induces a ranking of the candidates in $\mathcal{C}_{\mathrm{I}}(\tau)$, giving per-component ranks $r_{\mathrm{wr}}, r_{\mathrm{tr}}, r_{\mathrm{rot}}, r_{\mathrm{ft}} \in \{1, \dots, K_{\mathrm{I}}\}$, and the fused score $\rho$ of the main text sums them with equal weight,

\begin{equation}
\rho = r_{\mathrm{wr}} + r_{\mathrm{tr}} + r_{\mathrm{rot}} + r_{\mathrm{ft}} ,
\end{equation}

and targets are ordered by increasing $\rho$ (smaller is better).
We then deduplicate the sorted list \emph{by source trajectory}: scanning from the best-ranked target, we keep the first candidate of each source episode and discard its later duplicates, so that no source contributes more than once.
The first $K_{\mathrm{II}} = 100$ entries of the deduplicated list form the mined subset contributed by $\tau$,

\begin{equation}
\mathcal{C}_{\mathrm{II}}(\tau) \subseteq \mathcal{C}_{\mathrm{I}}(\tau),
\qquad \bigl|\mathcal{C}_{\mathrm{II}}(\tau)\bigr| \le K_{\mathrm{II}} .
\label{eq:dedup-shortlist}
\end{equation}

The set $\mathcal{C}_{\mathrm{II}}(\tau)$ has fewer than $K_{\mathrm{II}}$ entries only when $\mathcal{C}_{\mathrm{I}}(\tau)$ spans fewer than $K_{\mathrm{II}}$ source episodes.
The global mined subset $\mathcal{D}^{*}_{h}$ of the main text is the union of these per-anchor lists; this global deduplication keeps each target only once, even when several anchors retrieve it,

\begin{equation}
\mathcal{D}^{*}_{h} = \bigcup_{\tau \in \mathcal{D}_r} \mathcal{C}_{\mathrm{II}}(\tau).
\label{eq:global-union}
\end{equation}

The mined subset used for training is thus the output of the two selection stages: per-anchor recall, ranking, and source deduplication, followed by the global deduplication of Eq.~(\ref{eq:global-union}).

\subsection{Stage III: Re-ranking}
\label{app:rerank}

Re-ranking provides an embodiment-agnostic check based on optical flow.
Crucially, and consistent with the main text, this stage produces a \emph{verification and analysis signal only}: it does not alter the mined subset $\mathcal{D}^{*}_{h}$ of Eq.~(\ref{eq:global-union}) that is used for training.
It is applied to the top $K_{\mathrm{II}}$ candidates of each anchor rather than to the full pool, since dense flow is costly to compute.

\paragraph{Flow descriptor and re-ranking score.}
For an anchor--target pair we extract dense optical flow over a short clip around the matched frame.\footnote{Our implementation uses Farneb\"ack's algorithm, but the stage does not depend on this choice: any dense flow field computed identically for anchors and targets serves the same purpose.}
The flow is aggregated into a fixed-length, embodiment-agnostic clip-level descriptor $\bm{g} \in \mathbb{R}^{D}$.
The descriptor summarizes where in the frame motion occurs, in which directions it points, and how its magnitude evolves over the clip.
Because it is built from pixel-level motion, this descriptor is independent of hand morphology and of retargeting accuracy.
Writing $\bm{g}_c$ for the descriptor block of cue $c$, the re-ranking score between an anchor descriptor $\bm{g}$ and a target descriptor $\bm{g}'$ is a weighted sum of per-cue similarities,

\begin{equation}
s_{\mathrm{flow}}(\bm{g}, \bm{g}') = \sum_{c} \lambda_c \, s(\bm{g}_c, \bm{g}'_c),
\qquad \lambda_c \ge 0, \quad \sum_{c} \lambda_c = 1,
\end{equation}

where $c$ indexes the cues that make up the descriptor and $\lambda_c$ are their mixing weights.
We use one fixed setting of $\lambda_c$ that slightly favors spatial and directional agreement.
Since Stage~III is a training-free verification signal that does not enter the mined subset $\mathcal{D}^{*}_{h}$, neither the training set nor any success rate of \Ours depends on this choice.
Re-ranking the top-$K_{\mathrm{II}}$ candidates of each anchor by $s_{\mathrm{flow}}$ yields the verification ordering that the main text visualizes for three anchors (Fig.~\ref{fig:visualization}).

\subsection{Bimanual Handling}
\label{app:bimanual}

We collect here the design choices for the two-hand setting, which the main text does not spell out beyond the two shared segments $\bm{a}^{\mathrm{wr}}_t$ and $\bm{a}^{\mathrm{ft}}_t$ of Eq.~(\ref{eq:shared-action}).

\paragraph{Per-hand scoring with symmetric fusion.}
In Stages~I and~II, every score that depends on the hands is computed \emph{independently} for the left and right hands in the shared space, and the two per-hand values are combined symmetrically into a single bimanual score,

\begin{equation}
s_{\mathrm{bi}} = s^{\mathrm{L}} + s^{\mathrm{R}},
\end{equation}

where $s^{\mathrm{L}}$ and $s^{\mathrm{R}}$ are the same per-hand similarity evaluated on the left and the right hand.
This equal-weight combination treats the two hands symmetrically and introduces no hand-specific parameters; since only the ranking induced by the score is used, summing and averaging over the two
hands give identical rankings.
Eq.~(\ref{eq:pose-dist}) is the same rule applied to a distance, where smaller is better.

\paragraph{Mirror augmentation.}
Human demonstrations carry a handedness bias (e.g.\ a predominance of right-handed execution), which would leave the left-handed portion of the shared space under-populated.
To remove this bias, the preprocessing of the human data includes a random left--right mirror: a geometric reflection that swaps the left and right hands, reflects the wrist-local fingertip and wrist quantities accordingly, and flips the image horizontally.
Its only role is to make the handedness distribution of the shared space symmetric; the mirror replaces a demonstration rather than adding a second copy, so it does not change the size of the human pool.
The operation is purely geometric and preserves the shared representation of App.~\ref{app:representation}, so mirrored targets remain directly comparable to anchors under all similarities above.
One consequence is that a mined sample may be the mirrored realization of its source clip.
This mirror is applied when the human pool is built and is independent of any data augmentation used during VLA training.

\subsection{Retrieval Hyperparameters}
\label{app:retrieval-hparams}

Table~\ref{tab:retrieval-hparams} collects every hyperparameter of the cascade in one place.
These values were fixed once and used unchanged for all three tasks and all robot-data scales; we did not tune them against downstream success rate.
The cascade therefore has no task- or scale-specific configuration.

\begin{table}[t]
\centering
\caption{Retrieval hyperparameters of the SiMDex cascade. One configuration is used for all tasks
and all robot-data budgets; no value is tuned against downstream success rate.}
\label{tab:retrieval-hparams}
\begin{tabular}{llc}
\toprule
Stage & Setting & Value \\
\midrule
\multirow{5}{*}{I: Recall}
 & Pose descriptor $\hat{\bm{p}}$ & $\ell_2$-normalized, $\mathbb{R}^{15}$ per hand \\
 & Pose retrieval & Euclidean distance \\
 & Sentence embedding $\bm{e}$ & $\mathbb{R}^{384}$, cosine similarity \\
 & Rank-fusion constant $\kappa$ & $60$ \\
 & Fused candidate set $K_{\mathrm{I}}$ & $10{,}000$ \\
\midrule
\multirow{6}{*}{II: Ranking}
 & Wrist trajectory $\bm{F}_{\mathrm{wr}}$ & $31 \times 3$ \\
 & Fingertip trajectory $\bm{F}_{\mathrm{ft}}$ & $30 \times 15$ \\
 & Waveforms $\bm{w}_{\mathrm{tr}}, \bm{w}_{\mathrm{rot}}$ & $\mathbb{R}^{10}$ each \\
 & Component fusion $\rho$ & equal-weight rank sum \\
 & Source deduplication & one sample per source episode \\
 & Per-anchor shortlist $K_{\mathrm{II}}$ & $100$ \\
\midrule
\multirow{2}{*}{III: Re-ranking}
 & Applied to & top $K_{\mathrm{II}}$ candidates per anchor \\
 & Effect on training set & none (verification only) \\
\bottomrule
\end{tabular}
\end{table}

\section{Data and the Human Pool}
\label{app:data}

This appendix specifies the two data sources of Sec.~\ref{sec:implementation}: the human pool that \Ours mines and the robot demonstrations that serve as anchors.
We state how each corpus is turned into the shared representation of Sec.~\ref{sec:representation}, how large each is, and how the human pool is filtered before mining.

\subsection{From Egocentric Video to the Shared Descriptor}
\label{app:data:preproc}

\paragraph{Source.}
The human pool is derived from EgoDex~\citep{hoque:iclr26}, a publicly released egocentric manipulation dataset of $829$ hours over $194$ tabletop tasks, distributed as $338{,}000$ episodes and roughly $90$M frames recorded at $30$\,fps.
Each frame carries the 3D position and orientation of the upper-body joints, including $25$ joints per hand, together with a per-frame tracking confidence; each episode carries a natural-language description.
Two properties make the corpus usable as a retrieval target for our purpose.
First, the hand transforms are given per frame rather than per clip, which is what makes fine-grained motion comparison possible.
Second, every episode already comes with language, so the recall stage of Sec.~\ref{sec:mining} needs no additional annotation.

\paragraph{Preprocessing.}
We apply five operators, in this order, before any sample enters the pool.
\emph{(i) Ego-motion removal.} Wrist and fingertip transforms are re-expressed in a body-centered frame.
\emph{(ii) Validity filtering.} Frames with absent or low-confidence hand tracking are discarded, as are frames whose wrist speed is out of range.
\emph{(iii) Temporal smoothing.} The retained trajectories are smoothed along time.
\emph{(iv) Wrist-local retargeting.} They are then mapped onto the shared descriptor of
Sec.~\ref{sec:representation}.
\emph{(v) Handedness mirroring.} A random left--right mirror is applied as specified in
App.~\ref{app:bimanual}.

\paragraph{Slicing.}
Each surviving episode is sliced into frame-level samples at unit stride.
A sample pairs the observation and the language instruction at the current frame with a future target of $H=30$ steps in the shared action space---that is, one sample covers approximately one second of motion at $30$\,fps.
This is the unit of every sample count in this paper, for both sources (App.~\ref{app:data:unit}).

\subsection{Quality Filtering for Post-Training}
\label{app:data:filter}

\paragraph{Why the bar is higher here than in pre-training.}
Human video plays different roles in the two regimes.
In large-scale pre-training it shapes the backbone, and its sheer volume makes the result insensitive to any individual sample.
In post-training it is mixed with robot demonstrations at a $1{:}1$ ratio and supervises the same action head through the same loss, so a noisy human sample competes directly with a real robot demonstration for gradient.
We therefore hold the human data to a stricter standard when building the post-training pool, and it is this filtered pool---not the raw corpus---that both \Ours and the baseline draw from, so the stricter bar applies to both alike.
The filter uses only tracking quality and motion plausibility, not activity category; Table~\ref{tab:pool} lists the retained pool.

\subsection{Pool Statistics}
\label{app:data:pool}

Table~\ref{tab:pool} reports the provenance and size of the pool.
This pool is the human target set $\mathcal{D}_h$ over which the cascade of App.~\ref{app:method} operates.
Both methods see it in full: \Ours mines $\mathcal{D}_h$, and the baseline draws its human samples at random from the same $\mathcal{D}_h$ throughout training.
The comparison of Sec.~\ref{sec:main_results} therefore isolates selection alone: the model, schedule and data budget are identical.

\begin{table}[t]
\centering
\caption{Provenance and size of the human pool. The left column reports the public EgoDex release~\citep{hoque:iclr26}; the right column reports the pool obtained after the preprocessing and quality filtering of Apps.~\ref{app:data:preproc}--\ref{app:data:filter}, which \Ours mines and from which the baseline draws its human samples at random. Pool episodes are counted if they contribute at least one sample.}
\label{tab:pool}
\small
\begin{tabular}{lrr}
\toprule
 & EgoDex release & Post-training pool \\
\midrule
Duration            & $829$\,h        & ${\sim}297$\,h \\
Episodes            & $338{,}000$     & $164{,}959$ \ ($48.8\%$) \\
Frame-level samples & ${\sim}90$M     & $32{,}034{,}551$ \ (${\sim}36\%$) \\
Samples per episode & ${\sim}265$     & ${\sim}194$ \\
Tabletop tasks      & $194$           & --- \\
\bottomrule
\end{tabular}
\end{table}

\paragraph{Scale relative to the robot data.}
The human pool $\mathcal{D}_h$ is $23.7\times$ the size of the default robot-data budget: $32.0$M frame-level samples against $1.35$M (App.~\ref{app:data:robot}).
The subset \Ours retrieves from it, ${\sim}1.49$M samples, is $4.66\%$ of the pool and ${\sim}1.10\times$ the robot budget.
Its size is an outcome of the cascade: the per-anchor top-$K_{\mathrm{II}}$ shortlist followed by global deduplication.
Those parameters were fixed once and never tuned against downstream success (App.~\ref{app:retrieval-hparams}), so this size was not chosen to hit a target fraction, and the baseline sees the same number of human samples, drawn at random.

\subsection{Robot Demonstrations}
\label{app:data:robot}

\paragraph{Collection.}
Robot demonstrations are collected by bimanual teleoperation of a two-arm, two-hand platform by trained operators in a controlled laboratory setting, within a single application scenario (industrial assembly) covering the three tasks of Sec.~\ref{sec:tasks}.
Episodes are recorded at $30$\,fps, and the policy takes a single egocentric view as its visual input, from the platform's head-mounted camera for robot samples and from the wearer's headset for human samples.
Keeping the observation single-view and egocentric on both sides is what makes the two sources interchangeable at the model input.
Mixing the two therefore needs no view selection and no cross-view alignment.

\paragraph{Scale.}
Wrist poses, fingertip positions and joint readings are recovered by forward kinematics.
They pass through the same descriptor and the same unit-stride, $H=30$ slicing as the human data.
From the raw collection we curate a high-quality subset of ${\sim}1.35$M frame-level samples, or ${\sim}12.4$ hours.
The raw collection is about twice the size of this curated subset.
The additional material comes largely from early teleoperation sessions, when the operators were still building proficiency with the platform.
Execution in those sessions is less fluent, and mistakes and interruptions are more frequent.
This material is what the $2\times$ budget of Sec.~\ref{sec:scaling} draws on; the $1\times$ budget uses the curated subset alone.

\subsection{Sample Unit and Normalization}
\label{app:data:unit}

A sample is a single frame paired with its language instruction and a $30$-step future target in the shared action space of Sec.~\ref{sec:representation}.
Both corpora are recorded at $30$\,fps, sliced at unit stride with the same horizon, and mapped through the same retargeting operator, so one human sample and one robot sample cover the same amount of time.
In format, they differ only in which segments of the action vector carry supervision and in the joint slots of the state, which are zero-filled for human samples (Sec.~\ref{sec:training}).
All data budgets, including the $1{:}1$ mixture and the baseline's equal human-data budget, are counted in samples; our hour figures are $30$\,fps conversions, for intuition only.

\paragraph{Normalization.}
Action targets are normalized per source before training, using precomputed mean and standard deviation statistics.
The same statistics are used for \Ours, the baseline and every scaling configuration.
The relative contribution of the two streams is therefore set by the mixture ratio, not by the numerical scale of either source, since normalization removes that scale.

\section{What Gets Mined}
\label{app:mined}

\begin{figure}[p]
\centering
\includegraphics[width=\linewidth]{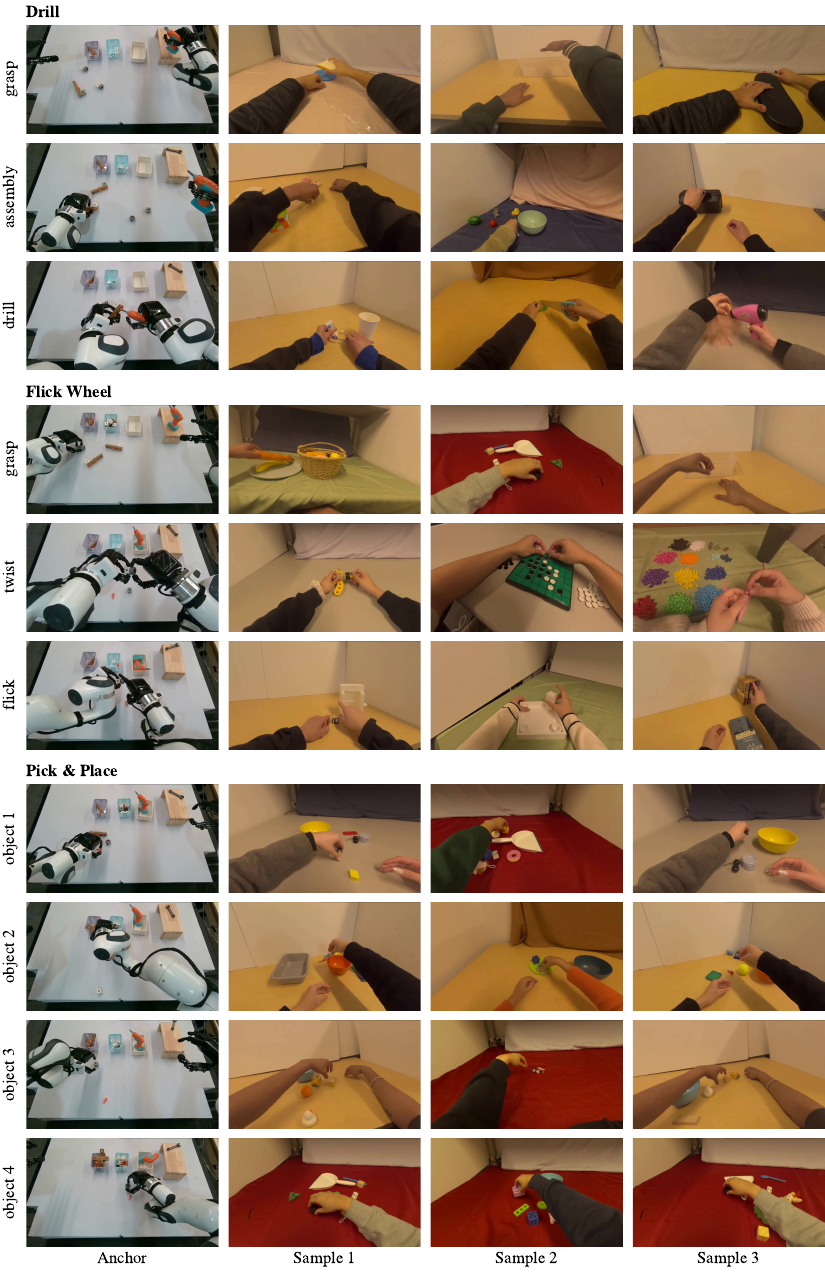}
\caption{\textbf{Every sub-task, and what is mined for it.} Rows follow the three tasks and their stages; the anchor comes first, then three mined human samples, each from a different source episode.}
\label{fig:mined-gallery}
\end{figure}

The main paper reports how much the mined data helps; this appendix reports what it \emph{is}.
The examples in the figures of this appendix are drawn from the mined subset used for training.
Stage~III does not alter that subset (Sec.~\ref{sec:mining}) and serves only as a check on it.

\subsection{What the Mined Set Covers}
\label{app:mined-coverage}

Fig.~\ref{fig:mined-gallery} shows all ten robot sub-tasks, covering every stage of every task.
The first cell of each row is the robot anchor; the three that follow are human samples mined for it.
No sub-task is left out.

\begin{figure}[t]
\centering
\includegraphics[width=\linewidth]{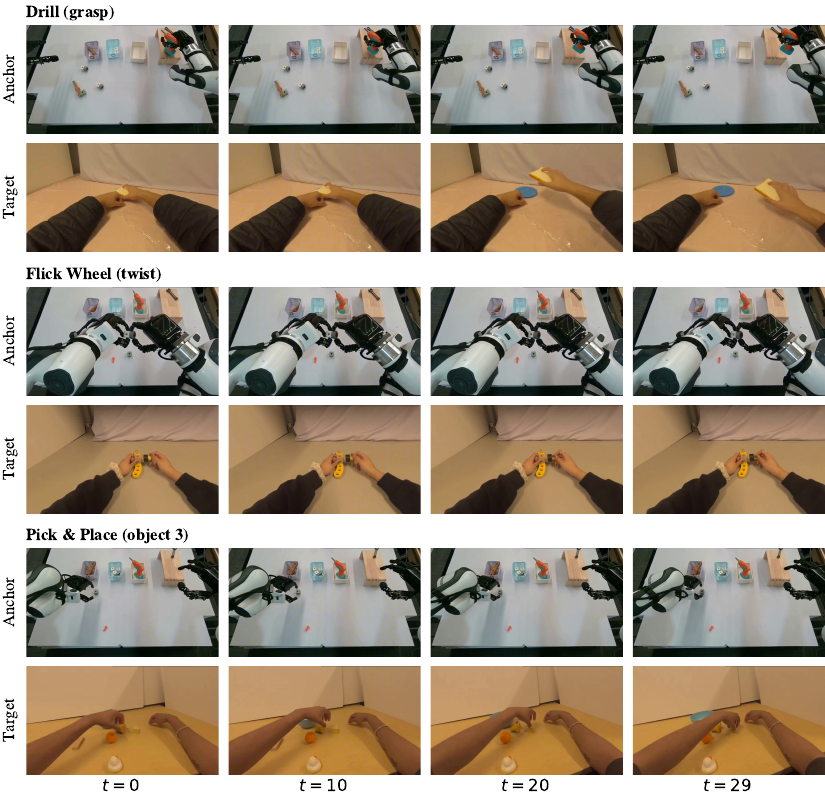}
\caption{\textbf{Anchors and the human samples mined for them.} One sub-task per task, at four positions of the $H=30$ horizon; within each block, anchor and target use the same frame indices.}
\label{fig:mined-pairs}
\end{figure}

\paragraph{Diversity of the mined set.}
Table~\ref{tab:mined-diversity} compares the mined set with GR-Dexter*, a fixed random subset of the same size drawn once from the full pool (scores in Table~\ref{tab:main} and Fig.~\ref{fig:scaling}).
The mined set draws on about half as many source episodes as GR-Dexter*, as expected of task-targeted mining, yet it covers nearly as many task instructions (about $2{,}100$ vs.\ $2{,}400$).
No episode or task instruction dominates: the largest episode holds $0.08\%$ of the set and the most frequent task instruction $2.0\%$.
For each anchor, \Ours keeps at most one sample per source episode (App.~\ref{app:ranking}), so a single episode contributes several samples only when many different anchors pick it.

\begin{table}[ht]
\centering
\caption{\textbf{Diversity of the mined set.} Percentages are shares of each set's $\sim$1.49M samples. Instructions are EgoDex's episode-level descriptions of task and scene, finer than its $194$ tasks.}
\label{tab:mined-diversity}
\small
\begin{tabular}{lcc}
\toprule
 & SiMDex (Ours) & GR-Dexter* \\
\midrule
EgoDex source episodes         & ${\sim}63{,}000$              & ${\sim}124{,}000$ \\
Largest single episode         & $1{,}171$ samples ($0.08\%$)  & $468$ samples ($0.03\%$) \\
Distinct task instructions     & about $2{,}100$               & about $2{,}400$ \\
Most frequent task instruction & $30{,}046$ samples ($2.0\%$)  & $13{,}779$ samples ($0.9\%$) \\
\bottomrule
\end{tabular}
\end{table}

\subsection{What the Match Is Based On}
\label{app:mined-examples}

Fig.~\ref{fig:mined-pairs} takes three sub-tasks, one per task, and shows the anchor and its mined sample as four frames each.
The four frames are four positions of the $H=30$ horizon, and the anchor and the sample use the same frame indices.
The useful thing to notice is what the pairs do \emph{not} share: environment, objects and hands all differ, and none of the clips is a recording of the robot's task.
What they share is the part the policy is supervised on: the wrist motion and the fingertip configuration over the horizon.
This is the intended behavior of a morphology-agnostic action space: matches are judged on motion structure, not on appearance.
It is also why visual verification (Stage~III) is informative rather than redundant: pixel-level motion carries information that the action-space stages never see.

\begin{figure}[t]
\centering
\includegraphics[width=\linewidth]{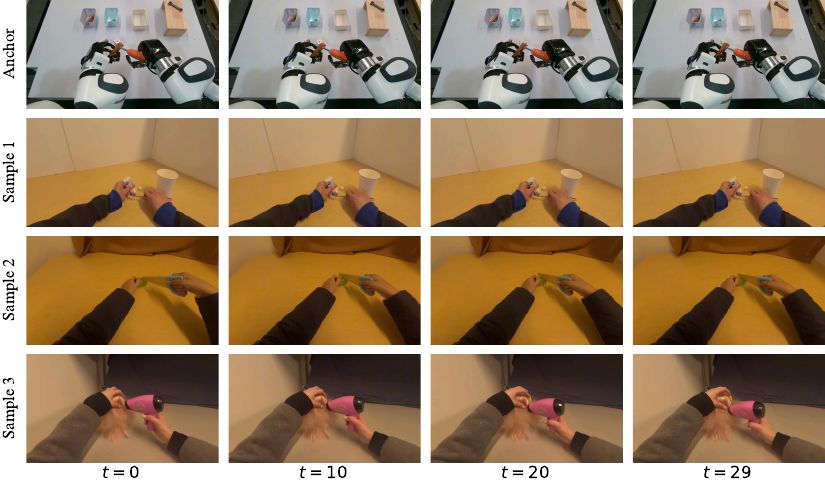}
\caption{\textbf{What is mined for the drilling stage of \emph{Drill}.} Below the anchor: a tube, a saw and a hair dryer, shown at the same four positions of the $H=30$ horizon. None of them is drilling.}
\label{fig:mined-drill}
\end{figure}

\subsection{How Retrieval Fails}
\label{app:mined-failures}

Reporting what the cascade retrieves would be incomplete without the cases where it retrieves the wrong thing.
Our similarity operates in a purely kinematic action space: wrist $6$D pose and fingertip positions.
It does not capture contact forces, object state or interaction semantics, and the errors below follow from this.
Three types of error recur; we give an example of each.

\textbf{Language-relevant but action-irrelevant.} Recall combines initial hand pose with instruction semantics, and the language side can carry a candidate on an object noun alone.
In the \emph{Flick Wheel} row of Fig.~\ref{fig:visualization}, the Recall result contains a wheel-like object, but the hand merely grasps it on a plane.
The word matches and the motion does not; Ranking addresses this by comparing motion.

\textbf{Similar motion, different interaction.} A candidate can match the anchor's translation and rotation waveforms while the hand is doing something else entirely.
The same pinch-and-turn rhythm appears in unscrewing, in wiping, and in rotating an object to inspect it.
Ranking narrows this considerably, since the fingertip trajectories have to agree as well.
But hands in distant positions can still score highly when their per-frame values happen to align, as the Ranking column of Fig.~\ref{fig:visualization} shows.

\textbf{The right tool in hand, the wrong interaction.} For the drilling stage of \emph{Drill}, the mined clips show a tube being squeezed, a saw, and a hair dryer (Fig.~\ref{fig:mined-drill}), and none of the EgoDex task labels of the clips mined for \emph{Drill} corresponds to drilling.
Holding a hand-held appliance and holding a drill are nearly identical in wrist pose and fingertip configuration.
The difference lies in contact, which is outside the action space, and in motion amplitude: the mined clips are oscillatory while the robot's drilling motion is nearly static, a difference the action-space scores largely discount, since they are built on cosine similarities and magnitude-normalized waveforms.
Nothing in the ranking marks such candidates as wrong, because on every signal it observes they look like correct matches.

The re-ranking stage of Sec.~\ref{sec:mining} targets exactly these cases: optical flow is computed on pixels, independent of the retargeting that both preceding stages depend on, so a candidate that ranks first in the action space can be demoted by where and how its motion actually occurs in the frame.
It reorders the same candidates that form the mined subset, so the samples it ranks first are already in the training data.
For the cost reason given in Sec.~\ref{sec:mining}, re-ranking in this work checks the mined subset rather than filtering it. Used as a filter on the same cascade output, it does not change the overall success rate significantly (61.9 vs.\ 58.3 without it, $p = 0.73$, Sec.~\ref{sec:stage-ablation}).
\newpage

\section{Experimental Setup}
\label{app:setup}

This appendix expands the implementation details of Sec.~\ref{sec:implementation}.
\Ours{} changes only the \emph{source} of the human data; the base model, the training procedure, and every hyperparameter are left untouched.
At every robot-data budget, \Ours{} and the baseline are therefore trained under identical settings.
The only variable in any comparison is thus which human samples are used.

\subsection{Base Model}
\label{app:setup:model}

Our base model is the $\pi_0$-like flow-matching VLA for dexterous manipulation introduced in Sec.~\ref{sec:implementation}; the baseline uses the identical model.
We summarize here only the components needed to interpret our experiments and to make the cross-embodiment supervision of Sec.~\ref{sec:training} precise; the remaining components are those of the base model, and \Ours{} modifies none of them.

\paragraph{Inputs and modules.}
At each timestep the model receives a single camera view $\bm{I}_t$, a language instruction $l$, and a proprioceptive state $\bm{s}_t \in \mathbb{R}^{86}$.
A vision-language backbone encodes the observation and the instruction.
A state adapter projects $\bm{s}_t$ into the backbone hidden space.
An action transformer, conditioned on the backbone output, then denoises a noisy action chunk into the predicted chunk $\bm{a}_{t:t+H} \in \mathbb{R}^{H \times 88}$ over the horizon $H = 30$.
The exact composition of the $88$ action dimensions and the $86$ state dimensions is given in App.~\ref{app:setup:layout}.
As stated in Sec.~\ref{sec:training}, this path is shared by robot and human samples; it is not conditioned on the source of a sample.
Human samples differ only in which state entries are populated and which action entries are supervised.

\paragraph{Training objective.}
Training uses the masked flow-matching objective of Eq.~(\ref{eq:loss}) in the main text.
Given a ground-truth action chunk $\bm{a} \in \mathbb{R}^{H \times 88}$, we sample noise $\bm{\epsilon} \in \mathbb{R}^{H \times 88}$ with entries drawn i.i.d.\ from $\mathcal{N}(0,1)$ and a timestep $\sigma \in [0,1]$, form the interpolated state $\bm{x}_\sigma$ together with its flow target $\bm{u}_\sigma$, and regress the predicted velocity $\hat{\bm{u}}_\sigma = f_\theta(\bm{x}_\sigma, \bm{I}_t, l, \bm{s}_t)$ under the masked mean squared error of Eq.~(\ref{eq:loss}).
The timestep $\sigma$ is not drawn uniformly from $[0,1]$; instead, we sample

\begin{equation}
v \sim \mathrm{Beta}(1.5,\, 1.0), \qquad \sigma = 0.999\,(1-v),
\label{eq:sigma}
\end{equation}

where $v \in [0,1]$ is an auxiliary scalar used only to generate the timestep $\sigma \in [0,1]$.
The model regresses the flow target rather than the clean action or the noise itself.
The action loss is the only objective active in all runs reported in this paper.
The backbone's language-modeling loss and the optional auxiliary losses are disabled throughout, for \Ours{} and the baseline alike.

\paragraph{What is held fixed.}
The architecture, the model capacity, the input and output dimensions, the optimizer, the schedule, and the number of training steps are identical across all compared methods.
\Ours requires only a backbone that reads a shared action layout with a per-dimension mask.

\subsection{Training Hyperparameters}
\label{app:setup:hparams}

All models---\Ours{}, the baseline, and every scaling configuration---are trained with the hyperparameters of Table~\ref{tab:hparams}.
The sole difference between \Ours{} and the baseline is which human samples are used.
We emphasize that none of these values were tuned per method: a single configuration, including the step count at $0.25\times$, is used for every run reported in this paper.

\begin{table}[t]
\centering
\caption{Training hyperparameters, shared by \Ours{}, the baseline, and all scaling configurations. Unlisted model and optimizer settings are also shared; no value was tuned per method.}
\label{tab:hparams}
\begin{tabular}{ll}
\toprule
Setting & Value \\
\midrule
Optimizer                       & AdamW \\                        %
Weight decay                    & $0.1$ \\                        %
Peak learning rate              & $3 \times 10^{-5}$ \\           %
LR schedule                     & cosine, $5\%$ linear warmup \\  %
Minimum learning rate           & $1 \times 10^{-6}$ \\           %
Gradient clipping               & enabled \\                      %
Mixed precision                 & bf16 \\                         %
\midrule
Training objective              & masked MSE (flow matching) \\   %
Noise distribution              & entries i.i.d.\ $\mathcal{N}(0,1)$ \\ %
Timestep sampling               & $\sigma = 0.999\,(1-v)$, $v \sim \mathrm{Beta}(1.5, 1.0)$ \\ %
Regression target               & flow target (velocity) \\       %
Supervised dims (robot / human) & $88 / 42$ \\                    %
Action horizon $H$              & $30$ \\                         %
\midrule
Training steps                  & $40$K ($20$K for the $0.25\times$ scale) \\ %
Robot\,:\,human mixture         & $1{:}1$ (equal-weight sampling) \\          %
Random seed                     & $42$ \\                                     %
\bottomrule
\end{tabular}
\end{table}

\paragraph{Two independently weighted data streams.}
The robot demonstrations and the mined human subset are not merged offline into a single dataset.
They are registered as two separate sources in the training pipeline and multiplexed by source weight during loading, so that the two contribute in a $1{:}1$ ratio in expectation rather than through a hard per-batch quota.
Because the mixture is governed by source weights and not by dataset size, the ratio is unaffected by how many samples each source happens to contain.
Action targets are normalized per source before training (App.~\ref{app:data:unit}).

\paragraph{Preprocessing and augmentation.}
The mined subset and the baseline's randomly sampled human data are consumed by the same pipeline: the mined data receives no additional filtering, reweighting or preprocessing of its own.
All data-augmentation settings are likewise identical for every method, so augmentation cannot explain any difference between \Ours{} and the baseline.

\subsection{Unified Action and State Layout}
\label{app:setup:layout}

The policy predicts an $88$-dimensional action chunk and consumes an $86$-dimensional state.
We give the exact composition, since it fixes which dimensions carry gradient for a human sample.

\paragraph{Action layout.}
Table~\ref{tab:action-layout} lists the four blocks of the action chunk.
Two of them---the bimanual wrist actions $\bm{a}^{\mathrm{wr}}_t$ and the bimanual fingertip targets $\bm{a}^{\mathrm{ft}}_t$---are expressed relative to the current wrist and are therefore defined for both embodiments; together they form the shared action $\bm{a}^{\mathrm{sh}}_t \in \mathbb{R}^{42}$ of App.~\ref{app:representation}.
The other two carry mechanism-level commands that have no human counterpart.

\begin{table}[t]
\centering
\caption{Composition of the $88$-dimensional action chunk. For human samples, the ``shared'' blocks are supervised, while the ``robot-only'' blocks are zero-filled and masked out.}
\label{tab:action-layout}
\setlength{\tabcolsep}{4pt} %
\begin{tabular}{lllcc}
\toprule
Segment & Symbol & Contents & Dim. & Human supervision \\
\midrule
Bimanual wrist actions    & $\bm{a}^{\mathrm{wr}}_t$ & $(\bm{d}^{\mathrm{L}}_t, \bm{d}^{\mathrm{R}}_t)$, $6$ per hand & $12$ & shared \\      %
Arm-joint actions         & $\bm{a}^{\mathrm{ar}}_t$ & joint commands, $7$ per arm                                    & $14$ & robot-only \\  %
Hand-joint commands       & $\bm{a}^{\mathrm{hd}}_t$ & joint commands, $16$ per hand                                  & $32$ & robot-only \\  %
Bimanual fingertip targets& $\bm{a}^{\mathrm{ft}}_t$ & $(\bm{f}^{\mathrm{L}}_t, \bm{f}^{\mathrm{R}}_t)$, $15$ per hand & $30$ & shared \\      %
\midrule
Total ($\bm{a}_t$)        &                          &                                                                & $88$ & $42$ supervised \\
\bottomrule
\end{tabular}
\end{table}

\paragraph{State layout.}
The state $\bm{s}_t \in \mathbb{R}^{86}$ concatenates the arm joint readings of both arms ($7$ per arm, $14$ in total), the hand joint readings of both hands ($21$ per hand, $42$ in total), and the current fingertip configurations $\bm{p}_t \in \mathbb{R}^{15}$ of both hands ($15$ per hand, $30$ in total).
Note that the hand joint \emph{readings} ($21$ per hand) and the hand joint \emph{commands} in the action chunk ($16$ per hand) live on different parameterizations, so the two counts need not match.
For a human sample the $56$ robot-specific state entries---arm and hand joints---have no counterpart and are zero-filled, while the $30$ fingertip entries are populated; the state adapter consumes the same $86$-dimensional vector in both cases.

\paragraph{Effective supervision for human samples.}
Under the mask $\bm{m}$ of Sec.~\ref{sec:training}, a human sample contributes gradient through the $12$ wrist dimensions and the $30$ fingertip dimensions, i.e.\ $42$ of the $88$ action dimensions.
The mask is applied per sample rather than per batch, so a single batch may freely mix the two sources.
Robot samples are supervised on all $88$ dimensions.

\subsection{Equal-Budget Comparison}
\label{app:setup:budget}

The comparison of Table~\ref{tab:main} is designed so that \emph{selection} is the only variable.

\paragraph{Matched human-data budget.}
\Ours{} trains on its ${\sim}1.49$M mined samples.
The baseline, GR-Dexter, draws human samples at random from the same ${\sim}32$M pool throughout training---the same pool, the same preprocessing, the same quality filtering---so it sees as many human samples as \Ours.
What is matched is the number of human samples seen, not a fixed subset size. An equal-size fixed random subset, GR-Dexter*, is reported separately (Table~\ref{tab:main}, Fig.~\ref{fig:scaling}).
Neither method sees human data that the other could not have seen; they differ only in which pool members they train on.

\paragraph{Everything else held fixed.}
The robot demonstrations, the model, the random seed, and every other entry of Table~\ref{tab:hparams} are identical between the two runs.
Both human subsets enter training through the same two-source pipeline described in App.~\ref{app:setup:hparams}, with the same source weight.
Apart from evaluation noise, any difference in success rate is therefore attributable to which human samples are used, and not to model capacity, training budget, data volume, or mixture composition.

\subsection{Data Scaling Ablation}
\label{app:scaling}

The scaling study of Sec.~\ref{sec:scaling} varies the robot-data budget while holding the human data fixed.
We describe how the budgets are built and the controls that make the four scales comparable.

\paragraph{The four robot-data budgets.}
The default budget ($1\times$, ${\sim}1.35$M frame-level samples, ${\sim}12.4$ hours) is a high-quality subset curated from the raw teleoperation collection, and is used everywhere else in the paper.
The $0.5\times$ (${\sim}6.2$h) and $0.25\times$ (${\sim}3.1$h) budgets are nested random subsets of it.
The $2\times$ budget (${\sim}24.9$h) is obtained by relaxing the quality filter over the \emph{same} raw collection---the same platform, the same tasks, the same operators---rather than by collecting new demonstrations.
We note the implication plainly: the largest budget trades some per-sample quality for quantity, so the $2\times$ column should be read as ``more robot data of mixed quality'' rather than as ``twice as much data of the $1\times$ standard''.
This does not affect the comparison at any given scale, since \Ours{} and the baseline consume exactly the same robot data at every budget.
It does, however, bear on comparisons across scales, such as \Ours{} at $0.5\times$ against the baseline at $2\times$, where about $6$ hours of curated robot data are set against about $25$ hours of mixed-quality robot data.

\paragraph{Nested subsampling.}
The smaller budgets are drawn from the $1\times$ set with a fixed random seed ($53$), so the scales are nested: every demonstration at $0.25\times$ is also present at $0.5\times$ and $1\times$.
The same subsets are used for \Ours{} and for the baseline, so both see identical robot data.

\paragraph{Fixed human data.}
The human data is identical across all scales.
\Ours{} reuses the same mined subset at every robot-data budget, and the baseline draws at random from the same full pool, so the scaling curves isolate the effect of robot-data quantity and do not confound it with a change in the human data.
In particular, the mined subset is \emph{not} re-mined at each scale; it is mined once against the $1\times$ anchors and reused unchanged.
At $0.5\times$ and $0.25\times$, the mined subset is thus retrieved with anchors from the full $1\times$ set, which includes demonstrations these scales do not train on.

\paragraph{Training-step control.}
All scales train for the $40$K steps of Table~\ref{tab:hparams}, with one deliberate exception: the $0.25\times$ scale is trained for $20$K steps to avoid overfitting on its very small robot-data budget.
This is a control against overfitting, not a tuned advantage, and it applies equally to both methods.

\paragraph{Evaluation rounds.}
Unlike the main results, which average over two rollout rounds ($2 \times 5 = 10$ trials), each scaling configuration is evaluated with a single rollout round.
The per-task values in the $1\times$ column of Fig.~\ref{fig:scaling} therefore differ from Table~\ref{tab:main}, as noted in the caption of Fig.~\ref{fig:scaling}.

\section{Evaluation Protocol}
\label{app:eval}

This section specifies how the real-robot rollouts of Sec.~\ref{sec:tasks} are run and scored.
Our goal is an evaluation fine-grained enough to credit partial progress, identical across all compared methods, and reproducible from the description alone.
Every element below---the task decomposition, the instruction set, the scoring rubric, the initialization, and the scoring personnel---is shared by \Ours and the baseline at every data budget, so the only variable across methods is the human-data source.

\subsection{Tasks, Language Instructions, and Scoring Rubric}
\label{app:eval:tasks}

The three evaluation tasks of Sec.~\ref{sec:tasks} are decomposed into ordered sub-tasks: three for Drill, three for Flick Wheel, and four for Pick \& Place, ten in total.
These sub-tasks are the unit of scoring in Table~\ref{tab:main}, and they are also the unit of language conditioning: each sub-task has one fixed instruction, and the policy is conditioned on the instruction of the sub-task currently in progress.
Table~\ref{tab:eval-instructions} lists all ten instructions verbatim with the observable steps on which each sub-task is scored.

\begin{table}[!ht]
\centering
\caption{\textbf{Evaluation tasks, language instructions, and scoring rubric.} The sub-tasks match the columns of Table~\ref{tab:main}; instructions are given verbatim as fed to the policy. Each sub-task is worth one point, split equally over the listed steps, and is complete when all of its steps are achieved.}
\label{tab:eval-instructions}
\small
\setlength{\tabcolsep}{4pt}
\begin{tabular}{@{}ll>{\raggedright\arraybackslash}p{4.6cm}>{\raggedright\arraybackslash}p{4.9cm}@{}}
\toprule
Task & Sub-task & Language instruction & Scored steps (points each) \\
\midrule
\multirow{3}{*}{\shortstack[l]{I: Drill\\(max 3)}}
 & Grasp  & \texttt{pick up the electric drill}
          & reach the drill; lift it and hold it stably in one hand; press the trigger ($\tfrac13$) \\
 & Assem. & \texttt{pick up the bar-to-cube assembly}
          & reach the assembly; grasp it; place it in alignment with the held drill ($\tfrac13$) \\
 & Drill  & \texttt{unscrew the cube with screw-holes from the bar-to-cube assembly}
          & bring the drill to the assembly; make contact; drive the screw-hole cube off the bar ($\tfrac13$) \\
\midrule
\multirow{3}{*}{\shortstack[l]{II: Flick Wheel\\(max 3)}}
 & Grasp  & \texttt{pick up the wheel-to-cube assembly}
          & reach the assembly; lift it and hold it stably ($\tfrac12$) \\
 & Twist  & \texttt{unscrew the bolted wheel from the wheel-to-cube assembly}
          & bring two fingers to the wheel; make contact; turn the wheel until the bolt is fully released ($\tfrac13$) \\
 & Flick  & \texttt{flick the wheel-to-cube assembly with index finger}
          & contact and flick the wheel with the index finger; the wheel comes off the assembly ($\tfrac12$) \\
\midrule
\multirow{4}{*}{\shortstack[l]{III: Pick \& Place\\(max 4)}}
 & Obj.1 & \texttt{put the plastic screw into the blue basket}
         & \multirow{4}{4.9cm}{\raggedright reach the named object; grasp it; place it inside the blue basket ($\tfrac13$)} \\
 & Obj.2 & \texttt{put the bolted double-strip bar into the blue basket} & \\
 & Obj.3 & \texttt{put the cube with screw-holes into the blue basket} & \\
 & Obj.4 & \texttt{put the bolted wheel into the blue basket} & \\
\bottomrule
\end{tabular}
\end{table}

\paragraph{Bimanual structure of Drill.}
Task~I is bimanual by construction: one hand acquires and holds the powered drill while the other acquires the workpiece and presents it to the tool, and only then is the tool actuated.
The three sub-tasks therefore test, in order, single-hand tool acquisition, second-hand workpiece acquisition with inter-hand alignment, and powered tool use under that alignment.
This is the property that makes Drill the most coordination-demanding of the three tasks, and it is why a failure to establish the alignment in the second sub-task leaves the third unreachable.

\paragraph{Instruction switching.}
Within a trial the policy is conditioned on one sub-task instruction at a time.
In Drill and Flick Wheel, the instruction advances to the next sub-task when the current one is complete and does not advance otherwise, so a trial that stalls remains conditioned on the sub-task it failed.
The one exception is the Drill Grasp: once the drill is held stably, a waiting period is allowed for the trigger press, and the instruction advances even if the trigger is not pressed within it.
In Pick \& Place, the instruction advances to the next object when the current one is placed or its attempt is terminated under the rules of Sec.~\ref{app:eval:trials}.
The switching decision is made by the evaluation team (Sec.~\ref{app:eval:independent}) under the same rules for every method, and the policy receives no other signal about task state.
This protocol is part of the harness and identical for both methods at every budget.

\subsection{Step-Based Sub-Task Scoring}
\label{app:eval:scoring}

Each sub-task is worth one point, divided equally among the observable steps of Table~\ref{tab:eval-instructions}: three steps of $\tfrac{1}{3}$ for most sub-tasks, and two steps of $\tfrac{1}{2}$ for the Grasp and Flick sub-tasks of Flick Wheel.
A sub-task's score is the sum of the steps achieved, taking values in $\{0, \tfrac{1}{3}, \tfrac{2}{3}, 1\}$ or $\{0, \tfrac{1}{2}, 1\}$, so that a rollout that gets most of the way through a contact-rich stage is distinguished from one that never engages the object.
On the score sheets, $\tfrac{1}{3}$ and $\tfrac{2}{3}$ are recorded as $0.33$ and $0.67$, and all reported scores use these recorded values.
Steps are credited only during their own sub-task: a Drill trigger press that occurs during a later sub-task, for example, does not retroactively earn the Grasp step.

\paragraph{Sequential dependence.}
Drill and Flick Wheel are sequential: their sub-tasks form a chain, and when a stage is not completed the subsequent stages cannot be reached and receive $0$.
The one step outside this chain is the trigger press of the Drill Grasp, which is scored but is not a precondition for assembly (Sec.~\ref{app:eval:tasks}).
Pick \& Place is different in kind: its four objects are attempted in the fixed order of Table~\ref{tab:eval-instructions}, but a failure on one object does not prevent the robot from attempting the next, and each object is therefore scored independently.
This distinction is visible in Table~\ref{tab:main}: the sub-task scores of Drill and Flick Wheel do not increase along the chain for either method, whereas those of Pick \& Place do not follow this pattern---the smallest object, the plastic screw (Obj.1), is the hardest object for both methods and scores below all three objects attempted after it.

\paragraph{From sub-task scores to reported numbers.}
A trial's score is the sum of its sub-task scores, so the per-trial maximum is $3$ for Drill, $3$ for Flick Wheel, and $4$ for Pick \& Place.
The Total column of Table~\ref{tab:main} is the mean of these per-trial sums over all trials, and the Success Rate is $\mathrm{Total}/\mathrm{max}\times 100\%$.
The Overall Success Rate is the unweighted mean of the three task success rates, each first rounded to one decimal place as printed, so it can be recomputed from the printed rates.

\subsection{Trial Structure, Initialization, and Reset}
\label{app:eval:trials}

\paragraph{Trials and rounds.}
Each task in Table~\ref{tab:main} is evaluated over $10$ trials in two rounds of five.
The five trials of a round use five distinct object layouts, indexed $1$--$5$; the second round repeats the same five layouts in the same order.
Each layout is therefore evaluated twice per method, and the two rounds differ only by repetition, not by initial conditions.
The two rounds are run as separate evaluation sessions, over which the reported standard deviations are computed (Sec.~\ref{app:eval:aggregation}).
The scaling study of Sec.~\ref{sec:scaling} uses a single round of five trials per configuration instead (App.~\ref{app:scaling}).

\paragraph{Controlled initialization.}
Initialization is governed by a fixed table-level mask: each of the five trial indices has one prescribed object layout, and before every trial the workspace is restored to that layout.
Restoration is verified against a third-person camera view rather than by eye, so a layout is reproduced from trial to trial and from method to method to within camera-verified placement error.
Initialization is controlled along two axes.
Across methods, trial index $i$ always uses layout $i$, so \Ours and the baseline start from identical initial conditions at every budget and are directly comparable trial by trial.
Across trial indices, the five layouts differ, so the five trials of a round probe five distinct object configurations rather than a single one.
The five layouts were chosen at random, and none of them reproduces a configuration that occurs in the training demonstrations: every evaluation trial therefore starts from an object configuration the policy has not seen during training.
The comparison is thus fair within each trial while still testing robustness to spatial variation---neither method can benefit from an easier draw, and neither is evaluated on only one layout.

\paragraph{Termination, retries, and reset.}
A trial ends when the task's final sub-task is completed.
When it is not completed, termination is decided by two rules.
The first is a stall allowance: a rollout that is still moving but is evidently not converging on the criterion---repeating the same approach, or advancing very slowly---is given roughly two further minutes, and if nothing changes it is terminated and scored at the progress reached.
The second is an attempt limit for sub-tasks that require picking an object off the table: the hand is allowed at most three attempts, and the sub-task is terminated after the third failure.
One attempt is one approach--close--lift cycle, counted from the moment the hand begins to descend toward the object until it either lifts the object clear of the table or ends up without it; a grasp that is established and then dropped consumes an attempt.
Retries are given by continued execution rather than by re-prompting: the instruction is not re-sent, and the policy is not reset or re-initialized within a trial.
No human intervention is permitted while a trial is running; the operators observe and score from outside the workspace and act only to halt the robot for safety.
Between trials the workspace is restored to the layout of the next trial index as described above.
The same termination, retry, and reset rules apply to every method and every trial.

\begin{figure}[t]
\centering
\includegraphics[width=\linewidth]{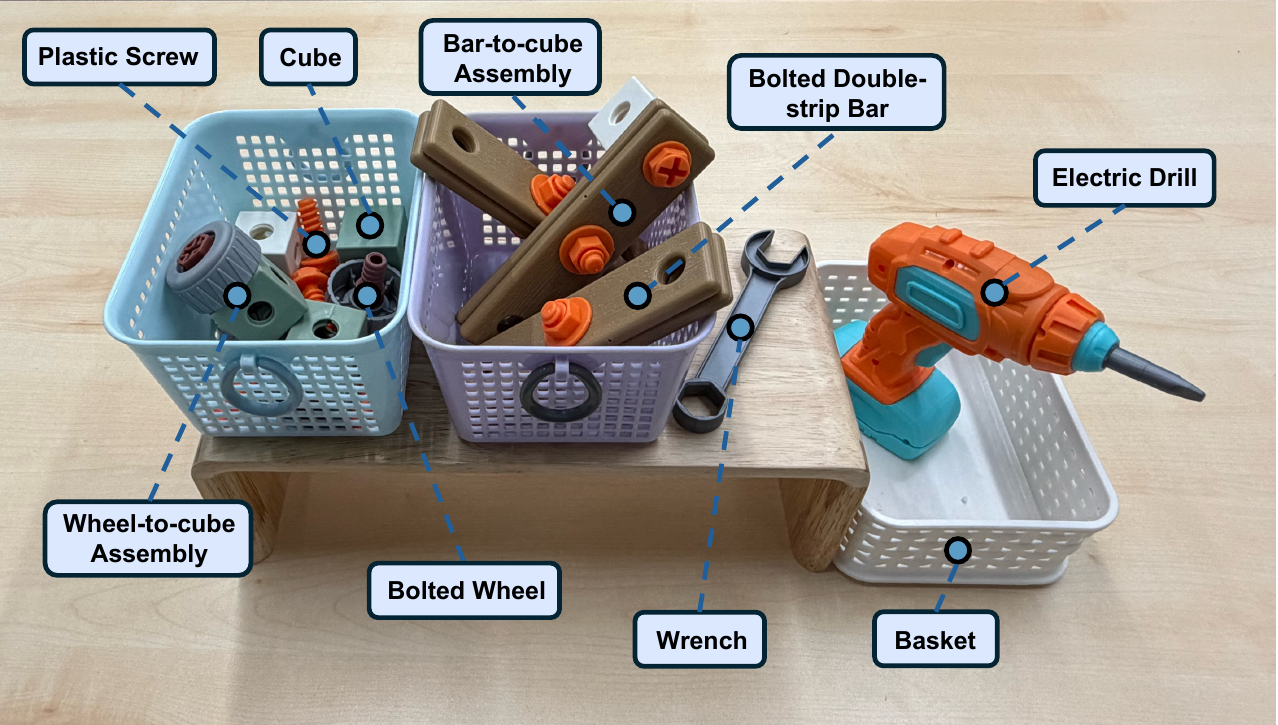}
\caption{\textbf{Objects and tools used in the real-robot evaluation.}
The drill is the tool; bars, wheels, cubes, and screws are manipulated parts; baskets are Task~III's placement targets.}
\label{fig:eval-tools}
\end{figure}

\subsection{Independent Execution and Scoring}
\label{app:eval:independent}

To remove any possibility of bias toward our own method, the rollouts are executed and scored by a dedicated team of trained operators that is independent of the authors.
The authors take no part in running the trials, in switching instructions during a trial, or in assigning scores.
The same team and the same rubric of Table~\ref{tab:eval-instructions} are applied to \Ours and the baseline at every budget, so no method receives preferential judgment.
The evaluation is moreover blind to method identity: the operators are not informed which method produced the policy under evaluation.
This matters in particular for the instruction switching of Sec.~\ref{app:eval:tasks}, whose timing rests on a human judgment: since the person making that judgment does not know whether \Ours or the baseline is running, the judgment cannot be systematically biased toward either.
Consistent with the trial protocol of App.~\ref{app:eval:trials}, no person entered the robot workspace during autonomous evaluation; trials were observed and scored from outside it by trained personnel who could halt the robot at any time.

\subsection{Aggregation and Variance Reporting}
\label{app:eval:aggregation}

\paragraph{What the reported numbers are.}
Each sub-task entry of Table~\ref{tab:main} is the mean of that sub-task's score over all $10$ trials of the task.
The reported $\pm$ is \emph{not} the standard deviation over the ten individual trials: it is the sample standard deviation over the two round-level means ($n=2$, i.e.\ $|a-b|/\sqrt{2}$), where each value is the mean of the five trials of one round.
It therefore measures the reproducibility of a full evaluation session rather than trial-to-trial spread, and with $n=2$ it should be read as an indication of session-level stability, not a confidence interval.
We report it this way because the two rounds are the unit at which the evaluation was repeated.
Because the two rounds repeat the same five layouts (Sec.~\ref{app:eval:trials}), this $\pm$ compares two evaluations under identical initial conditions: it isolates the run-to-run repeatability of the policy and hardware, and does not absorb layout variation.

\paragraph{Paired tests.}
The test of Sec.~\ref{sec:main_results} pairs the trials of Table~\ref{tab:main} by task, layout and round.
It is a two-sided exact permutation test, and the CI is a 95\% bootstrap interval of the mean paired difference.

\subsection{Objects, Tools, and Workspace}
\label{app:eval:tools}

Fig.~\ref{fig:eval-tools} shows the objects and tools used in the evaluation: a powered drill, bolted double-strip bars, wheels, cubes with screw holes, plastic screws, and the baskets that serve as placement targets.
The parts are industrial-assembly hardware with tight tolerances, which is what makes the tasks contact-rich: the screw-hole cube and the bolted wheel must be driven or turned off their mating parts rather than pulled, and the plastic screw is small enough that a grasp must be precise to succeed at all.
The identical object set, in the same physical condition, is used for every method and every trial; objects are not substituted or re-selected between methods.
Details of the robot platform and teleoperation hardware are omitted here and are not required to interpret the protocol: the evaluation is defined entirely by the tasks, language instructions, rubric, and layouts specified above.

\section{Additional Results}
\label{app:results}

This section reports the full numbers behind Figure~\ref{fig:scaling}, which the main text shows only as bar charts.
We first define the four robot-data budgets and give per-sub-task scores for every task, budget, and method, together with the equal-size random subset (GR-Dexter*) and the robot-only references, then add stage-entry counts, qualitative rollouts, and the multi-view recording setup.

\subsection{Complete Data-Scaling Results}
\label{sec:supp-scaling-full}

Table~\ref{tab:scale-def} defines the four robot-data budgets.
The mined human data is identical across all budgets, as are the model and every hyperparameter, so the only variable is the robot-data budget (App.~\ref{app:scaling}).
The one exception is that the smallest budget ($0.25\times$) is trained for 20K steps rather than 40K, to avoid overfitting on a very small robot-data budget; this applies equally to both methods.

\begin{table}[!ht]
\centering
\caption{\textbf{Robot-data budgets used in the scaling study.} Mined human data is fixed at ${\sim}1.49$M samples throughout. Samples are frame-level, scaled from the $1\times$ setting of Sec.~\ref{sec:implementation}.}
\label{tab:scale-def}
\begin{tabular}{lccc}
\toprule
Robot-data budget & Demonstrations & Frame-level samples & Training steps \\
\midrule
$0.25\times$ & ${\sim}3.1$\,h  & ${\approx}0.34$M & 20K \\
$0.5\times$  & ${\sim}6.2$\,h  & ${\approx}0.68$M & 40K \\
$1\times$    & ${\sim}12.4$\,h & ${\approx}1.35$M & 40K \\
$2\times$    & ${\sim}24.9$\,h & ${\approx}2.71$M & 40K \\
\bottomrule
\end{tabular}
\end{table}

Table~\ref{tab:scaling-full} gives the complete per-sub-task results.
Unlike Table~\ref{tab:main} in the main text, all scaling results come from a single rollout round of five trials per task, whereas Table~\ref{tab:main} averages two rounds (ten trials).
The $1\times$ column of Table~\ref{tab:scaling-full} therefore differs numerically from Table~\ref{tab:main}, as already noted in the caption of Figure~\ref{fig:scaling}.
Per-task success rate is the task's composite score divided by its maximum, and the overall success rate is the arithmetic mean of the three per-task success rates.

\begin{table}[p]
\centering
\footnotesize
\setlength{\tabcolsep}{4.5pt}
\caption{\textbf{Complete results behind Figure~\ref{fig:scaling}.} Sub-task scores are means over the five trials of a single rollout round (for $1\times$, the first round of Table~\ref{tab:main}). Total is the composite score; Success Rate $=$ Total\,/\,max\,$\times100\%$. Overall SR is the mean of the three per-task success rates. $\Delta =$ SiMDex $-$ GR-Dexter, computed from unrounded scores, so it may differ by one unit in the last digit from the difference of the printed entries; the success-rate deltas are the values printed in Figure~\ref{fig:scaling}.}
\label{tab:scaling-full}
\begin{tabular}{llcccccc}
\toprule
\multicolumn{8}{l}{\textbf{Task I: Drill} (max 3)} \\
\midrule
Budget & Method & Grasp & Assem. & Drill & & Total & SR(\%)$\uparrow$ \\
\midrule
\multirow{3}{*}{$0.25\times$} & GR-Dexter & 0.47 & 0.27 & 0.07 & & 0.80 & 26.6 \\
 & SiMDex (Ours) & \textbf{0.60} & \textbf{0.33} & \textbf{0.13} & & \textbf{1.07} & \textbf{35.5} \\
 & $\Delta$ & $+0.13$ & $+0.07$ & $+0.07$ & & $+0.27$ & $+8.9$ \\
\midrule
\multirow{3}{*}{$0.5\times$} & GR-Dexter & 0.47 & 0.40 & \textbf{0.20} & & 1.07 & 35.5 \\
 & SiMDex (Ours) & \textbf{0.67} & \textbf{0.67} & 0.13 & & \textbf{1.47} & \textbf{48.9} \\
 & $\Delta$ & $+0.20$ & $+0.27$ & $-0.07$ & & $+0.40$ & $+13.4$ \\
\midrule
\multirow{3}{*}{$1\times$} & GR-Dexter & \textbf{0.80} & \textbf{0.80} & \textbf{0.33} & & \textbf{1.93} & \textbf{64.5} \\
 & SiMDex (Ours) & 0.67 & 0.40 & 0.07 & & 1.13 & 37.7 \\
 & $\Delta$ & $-0.13$ & $-0.40$ & $-0.27$ & & $-0.80$ & $-26.7$ \\
\midrule
\multirow{3}{*}{$2\times$} & GR-Dexter & \textbf{1.00} & 0.87 & 0.20 & & \textbf{2.06} & \textbf{68.8} \\
 & SiMDex (Ours) & 0.74 & 0.87 & \textbf{0.26} & & 1.87 & 62.2 \\
 & $\Delta$ & $-0.26$ & $0.00$ & $+0.07$ & & $-0.20$ & $-6.6$ \\
\midrule[\heavyrulewidth]
\multicolumn{8}{l}{\textbf{Task II: Flick Wheel} (max 3)} \\
\midrule
Budget & Method & Grasp & Twist & Flick & & Total & SR(\%)$\uparrow$ \\
\midrule
\multirow{3}{*}{$0.25\times$} & GR-Dexter & 0.50 & 0.00 & 0.00 & & 0.50 & 16.7 \\
 & SiMDex (Ours) & \textbf{0.60} & \textbf{0.20} & \textbf{0.10} & & \textbf{0.90} & \textbf{30.0} \\
 & $\Delta$ & $+0.10$ & $+0.20$ & $+0.10$ & & $+0.40$ & $+13.3$ \\
\midrule
\multirow{3}{*}{$0.5\times$} & GR-Dexter & 0.70 & 0.27 & 0.10 & & 1.07 & 35.5 \\
 & SiMDex (Ours) & \textbf{0.80} & \textbf{0.47} & \textbf{0.20} & & \textbf{1.47} & \textbf{48.9} \\
 & $\Delta$ & $+0.10$ & $+0.20$ & $+0.10$ & & $+0.40$ & $+13.3$ \\
\midrule
\multirow{3}{*}{$1\times$} & GR-Dexter & 0.50 & 0.00 & 0.00 & & 0.50 & 16.7 \\
 & SiMDex (Ours) & \textbf{0.90} & \textbf{0.53} & \textbf{0.10} & & \textbf{1.53} & \textbf{51.1} \\
 & $\Delta$ & $+0.40$ & $+0.53$ & $+0.10$ & & $+1.03$ & $+34.5$ \\
\midrule
\multirow{3}{*}{$2\times$} & GR-Dexter & 0.60 & 0.07 & 0.00 & & 0.67 & 22.2 \\
 & SiMDex (Ours) & \textbf{0.70} & \textbf{0.27} & 0.00 & & \textbf{0.97} & \textbf{32.2} \\
 & $\Delta$ & $+0.10$ & $+0.20$ & $0.00$ & & $+0.30$ & $+10.0$ \\
\midrule[\heavyrulewidth]
\multicolumn{8}{l}{\textbf{Task III: Pick \& Place} (max 4)} \\
\midrule
Budget & Method & Obj.1 & Obj.2 & Obj.3 & Obj.4 & Total & SR(\%)$\uparrow$ \\
\midrule
\multirow{3}{*}{$0.25\times$} & GR-Dexter & 0.20 & \textbf{0.87} & \textbf{0.40} & \textbf{0.33} & \textbf{1.80} & \textbf{45.0} \\
 & SiMDex (Ours) & 0.20 & 0.80 & 0.40 & 0.33 & 1.73 & 43.2 \\
 & $\Delta$ & $0.00$ & $-0.07$ & $0.00$ & $0.00$ & $-0.07$ & $-1.8$ \\
\midrule
\multirow{3}{*}{$0.5\times$} & GR-Dexter & 0.33 & 0.73 & 0.40 & 0.60 & 2.06 & 51.5 \\
 & SiMDex (Ours) & \textbf{0.47} & \textbf{1.00} & \textbf{0.93} & \textbf{0.67} & \textbf{3.07} & \textbf{76.7} \\
 & $\Delta$ & $+0.13$ & $+0.27$ & $+0.54$ & $+0.07$ & $+1.01$ & $+25.2$ \\
\midrule
\multirow{3}{*}{$1\times$} & GR-Dexter & 0.26 & 0.60 & 0.60 & 0.73 & 2.19 & 54.8 \\
 & SiMDex (Ours) & \textbf{0.80} & \textbf{0.87} & \textbf{1.00} & \textbf{0.80} & \textbf{3.47} & \textbf{86.7} \\
 & $\Delta$ & $+0.54$ & $+0.27$ & $+0.40$ & $+0.07$ & $+1.28$ & $+31.9$ \\
\midrule
\multirow{3}{*}{$2\times$} & GR-Dexter & 0.26 & 0.87 & 0.53 & 0.80 & 2.46 & 61.6 \\
 & SiMDex (Ours) & \textbf{0.53} & \textbf{0.93} & \textbf{0.80} & 0.80 & \textbf{3.07} & \textbf{76.7} \\
 & $\Delta$ & $+0.27$ & $+0.07$ & $+0.27$ & $0.00$ & $+0.60$ & $+15.1$ \\
\midrule[\heavyrulewidth]
\multicolumn{8}{l}{\textbf{Overall Success Rate} (mean of the three per-task rates)} \\
\midrule
Budget & \multicolumn{3}{l}{GR-Dexter} & \multicolumn{2}{l}{SiMDex (Ours)} & & $\Delta$ \\
\midrule
$0.25\times$ & \multicolumn{3}{l}{29.4} & \multicolumn{2}{l}{\textbf{36.2}} & & $+6.8$ \\
$0.5\times$  & \multicolumn{3}{l}{40.8} & \multicolumn{2}{l}{\textbf{58.2}} & & $+17.3$ \\
$1\times$    & \multicolumn{3}{l}{45.3} & \multicolumn{2}{l}{\textbf{58.5}} & & $+13.2$ \\
$2\times$    & \multicolumn{3}{l}{50.9} & \multicolumn{2}{l}{\textbf{57.0}} & & $+6.2$ \\
\bottomrule
\end{tabular}
\end{table}

\paragraph{What the full table adds.}
The table makes three statements of the main text checkable cell by cell.
First, SiMDex's overall success rate exceeds the baseline at all four budgets, while the baseline's falls monotonically from 50.9\% at $2\times$ to 29.4\% at $0.25\times$.
Second, Flick Wheel benefits at all four budgets and Pick \& Place at all but the smallest.
Third, the Drill gap varies in sign; with one round of five trials per configuration, single-budget per-task differences are only indicative.

\begin{figure}[!t]
\centering
\includegraphics[width=\textwidth]{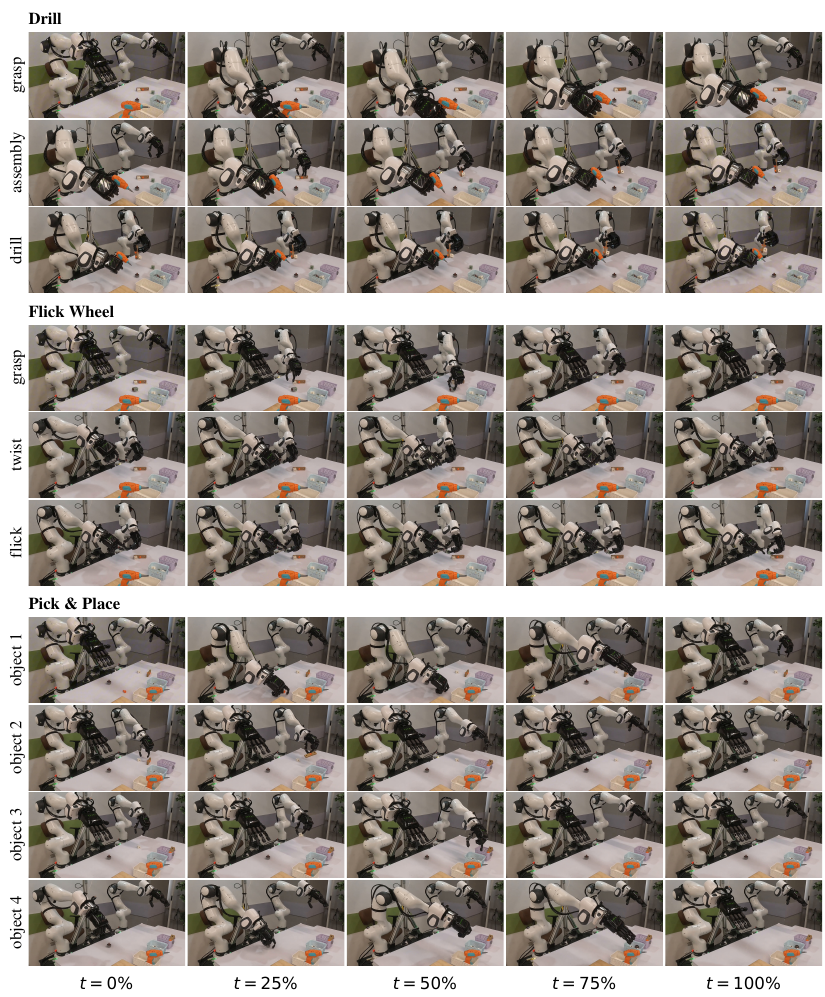}
\caption{\textbf{SiMDex rollouts on all ten evaluation sub-tasks.} One row per sub-task, with columns sampling progress within that sub-task ($t=0\%$ to $t=100\%$) rather than wall-clock time. Each row is a single rollout and not a restatement of the ten-trial means of Table~\ref{tab:main}; see Sec.~\ref{sec:supp-rollouts}.}
\label{fig:rollout-gallery}
\end{figure}

\paragraph{Equal-size random subset.}
GR-Dexter* is trained on a fixed random subset of the same size as the mined set: ${\sim}1.49$M samples drawn once from the full ${\sim}32$M pool, with the same preprocessing and quality filtering, and kept fixed thereafter.
At every budget it uses the same model, training settings and 1:1 mixture as \Ours.
It is evaluated with the same rubric and layouts (Table~\ref{tab:equal-size}).

\begin{table}[t]
\centering
\caption{\textbf{Equal-size random subset at every robot-data budget.} Success rate (\%) per task and robot-data budget. Each entry is the mean of five trials (one round). At $1\times$ it averages both rounds (ten trials), as in Table~\ref{tab:main}, whereas Figure~\ref{fig:scaling} and Table~\ref{tab:scaling-full} use the first round only. Overall is the mean of the three task success rates. Best overall success rate at each budget in \textbf{bold}.}
\label{tab:equal-size}
\small
\begin{tabular}{llcccc}
\toprule
Method & Task & $0.25\times$ & $0.5\times$ & $1\times$ & $2\times$ \\
\midrule
\multirow{4}{*}{GR-Dexter*} & Drill & 8.8 & 15.5 & 31.1 & 35.5 \\
 & Flick Wheel & 22.2 & 20.0 & 32.2 & 37.8 \\
 & Pick \& Place & 44.9 & 54.9 & 50.7 & 71.6 \\
 & Overall & 25.3 & 30.1 & 38.0 & 48.3 \\
\midrule
\multirow{4}{*}{GR-Dexter} & Drill & 26.6 & 35.5 & 64.5 & 68.8 \\
 & Flick Wheel & 16.7 & 35.5 & 24.5 & 22.2 \\
 & Pick \& Place & 45.0 & 51.5 & 54.0 & 61.6 \\
 & Overall & 29.4 & 40.8 & 47.7 & 50.9 \\
\midrule
\multirow{4}{*}{SiMDex (Ours)} & Drill & 35.5 & 48.9 & 54.5 & 62.2 \\
 & Flick Wheel & 30.0 & 48.9 & 47.2 & 32.2 \\
 & Pick \& Place & 43.2 & 76.7 & 83.3 & 76.7 \\
 & Overall & \textbf{36.2} & \textbf{58.2} & \textbf{61.7} & \textbf{57.0} \\
\bottomrule
\end{tabular}
\end{table}

The pooled tests combine the four budgets. Each pair matches two conditions on the same task, layout and budget, giving $60$ pairs. At $1\times$, where every condition has two rounds, the two trials on each layout are averaged first, so that every budget counts equally.
Each difference is the mean over matched pairs of unrounded trial scores, so it can differ by $0.1$ from the difference of the table entries. The $p$-values are from a two-sided paired permutation test, and CIs are $95\%$ bootstrap intervals.

\textbf{Against GR-Dexter*,} \Ours is higher at every budget. Pooled over the budgets, it is $+17.8$ higher overall ($p<0.0001$, 95\% CI $[+11.8, +24.1]$), $+27.5$ on Drill ($p<0.001$), $+11.5$ on Flick Wheel ($p=0.03$) and $+14.5$ on Pick \& Place ($p<0.01$).
\textbf{Against GR-Dexter,} \Ours is also higher at every budget. Pooled, it is $+11.1$ higher overall ($p<0.01$, 95\% CI $[+3.9, +18.2]$), $+14.9$ on Flick Wheel ($p<0.01$) and $+17.0$ on Pick \& Place ($p=0.001$), while the two are level on Drill ($+1.4$, $p=0.85$).
\textbf{Between the two random conditions,} drawing from the full pool (GR-Dexter) instead of the fixed subset (GR-Dexter*) adds $+6.8$ ($p=0.09$, not significant), mostly on Drill. Within this range of pool sizes, how samples are selected matters more than a larger random pool.

\paragraph{Per-task view: amount and coverage on Drill.}
Table~\ref{tab:drill-amount-coverage} averages the four robot-data budgets. When random human data come from the full pool (GR-Dexter) rather than the equal-size subset (GR-Dexter*), only Drill gains: $+26.1$ ($p<0.01$), against $-3.3$ on Flick Wheel and $-2.5$ on Pick \& Place (neither significant).
The pool does not cover drilling: none of the $194$ EgoDex tasks involves it, and the clips mined for the drilling stage show a tube, a saw and a hair dryer (App.~\ref{app:mined-failures}).
Both the amount and the coverage of human data therefore affect the success rate on Drill.

\begin{table}[t]
\centering
\caption{%
\textbf{Per-task success rate (\%) of the three human-data conditions, averaged over the four robot-data budgets.}
The budgets count equally: each entry is the plain mean of the four budget columns of Table~\ref{tab:equal-size}. Values in parentheses are for the $1\times$ budget alone and match Table~\ref{tab:main}.}
\label{tab:drill-amount-coverage}
\begin{tabular}{lccc}
\toprule
Method & Drill & Flick Wheel & Pick \& Place \\
\midrule
GR-Dexter*    & 22.7 (31.1) & 28.1 (32.2) & 55.5 (50.7) \\
GR-Dexter     & 48.9 (64.5) & 24.7 (24.5) & 53.0 (54.0) \\
SiMDex (Ours) & 50.3 (54.5) & 39.6 (47.2) & 70.0 (83.3) \\
\bottomrule
\end{tabular}
\end{table}

\paragraph{Robot-only references.}\phantomsection\label{app:robot-only}
As a reference, we also post-train policies on robot data alone: they start from the same pre-trained model as every other policy and are scored with the same rubric on the same layouts.
\emph{Robot only, same passes} sees the robot data as many times as the 1:1 mixed runs at the same budget.
\emph{Robot only, same training steps} uses the same training steps as \Ours at $1\times$. Since half of each mixed batch is human data, it sees the robot data twice as often as \Ours.

\begin{table}[!ht]
\centering
\caption{%
\textbf{Robot-only references across robot-data budgets.}
Success rate (\%) per task and budget. Each entry is the mean of five trials, or of ten where two rounds are averaged (\Baseline and \Ours at $1\times$, as in Table~\ref{tab:main}).
Overall is the mean of the three task scores shown.}
\label{tab:robot-only-budgets}
{\small\begin{tabular}{llcccc}
\toprule
Training data & Task & $0.25\times$ & $0.5\times$ & $1\times$ & $2\times$ \\
\midrule
\multirow{4}{*}{Robot only, same passes} & Drill & 11.0 & 11.0 & 42.2 & 24.3 \\
 & Flick Wheel & 16.7 & 40.1 & 32.3 & 30.0 \\
 & Pick \& Place & 39.8 & 53.2 & 56.6 & 66.6 \\
 & Overall & 22.5 & 34.8 & 43.7 & 40.3 \\
\midrule
\multirow{4}{*}{GR-Dexter} & Drill & 26.6 & 35.5 & 64.5 & 68.8 \\
 & Flick Wheel & 16.7 & 35.5 & 24.5 & 22.2 \\
 & Pick \& Place & 45.0 & 51.5 & 54.0 & 61.6 \\
 & Overall & 29.4 & 40.8 & 47.7 & 50.9 \\
\midrule
\multirow{4}{*}{SiMDex (Ours)} & Drill & 35.5 & 48.9 & 54.5 & 62.2 \\
 & Flick Wheel & 30.0 & 48.9 & 47.2 & 32.2 \\
 & Pick \& Place & 43.2 & 76.7 & 83.3 & 76.7 \\
 & Overall & \textbf{36.2} & \textbf{58.2} & \textbf{61.7} & \textbf{57.0} \\
\bottomrule
\end{tabular}\par}

\vspace{0.8em}
\caption{%
\textbf{Robot-only reference with the same training steps ($1\times$).}
Success rate (\%). All three rows average two rounds, and the \Baseline and \Ours rows are those of Table~\ref{tab:main}.}
\label{tab:robot-only-steps}
{\small\begin{tabular}{lcccc}
\toprule
Training data ($1\times$) & Drill & Flick Wheel & Pick \& Place & Overall \\
\midrule
Robot only, same training steps & 51.2 & 53.9 & 53.2 & 52.8 \\
GR-Dexter & 64.5 & 24.5 & 54.0 & 47.7 \\
SiMDex (Ours) & 54.5 & 47.2 & 83.3 & \textbf{61.7} \\
\bottomrule
\end{tabular}\par}
\end{table}

With the same passes, \Ours scores above the robot-only policies at every budget, by $+18.0$ on average over the four budgets (Table~\ref{tab:robot-only-budgets}).
With the same training steps ($1\times$), the robot-only policy reaches 52.8 overall, and \Ours (61.7) is still higher, by $+8.9$ (Table~\ref{tab:robot-only-steps}).

\subsection{Sub-task \texorpdfstring{Stage-Entry Counts}{Stage-Entry Counts}}
\label{sec:supp-cond-pass}

In Drill and Flick Wheel the sub-tasks form a chain (App.~\ref{app:eval:scoring}): a stage is entered only after the previous one is complete, and in Drill, Assembly also starts once the drill is held stably without the trigger press.
Raw sub-task scores therefore conflate two different things: how often the policy reaches a stage at all, and how well it does once it is there.
To separate the two, Table~\ref{tab:cond-pass} gives, for each stage and method, how many of the ten trials at $1\times$ enter the stage, how many complete it, and the mean progress of those that enter it.
Pick \& Place is excluded because its four objects are scored separately rather than forming a chain.
All counts come directly from the same trials and sub-task scores as Table~\ref{tab:main} and introduce no new data.

\begin{table}[t]
\centering
\caption{\textbf{Stage-entry counts on the two sequential tasks at $1\times$ (ten trials per method: two rounds $\times$ five layouts).} Completed\,/\,entered gives how many trials complete a stage out of those that enter it. Progress is the mean stage score of the trials that enter it. All trials are those of Table~\ref{tab:main}.}
\label{tab:cond-pass}
\begin{tabular}{llcccc}
\toprule
 & & \multicolumn{2}{c}{GR-Dexter} & \multicolumn{2}{c}{SiMDex (Ours)} \\
\cmidrule(lr){3-4}\cmidrule(lr){5-6}
Task & Stage & Completed\,/\,entered & Progress & Completed\,/\,entered & Progress \\
\midrule
\multirow{3}{*}{Drill} & Grasp & 6\,/\,10 & 0.80 & 4\,/\,10 & 0.73 \\
 & Assembly & 8\,/\,8 & 1.00 & 6\,/\,8 & 0.79 \\
 & Drilling & 0\,/\,8 & 0.42 & 0\,/\,6 & 0.45 \\
\midrule
\multirow{3}{*}{Flick Wheel} & Grasp & 2\,/\,10 & 0.60 & 7\,/\,10 & 0.85 \\
 & Twist & 0\,/\,2 & 0.67 & 3\,/\,7 & 0.67 \\
 & Flick & 0\,/\,0 & -- & 0\,/\,3 & 0.33 \\
\bottomrule
\end{tabular}
\end{table}

\paragraph{On Flick Wheel, SiMDex completes more stages.}
SiMDex completes Grasp in 7 of 10 trials and Twist in 3, against 2 and 0 for GR-Dexter, so only SiMDex reaches the final Flick stage (3 vs.\ 0 trials).
Trials that reach Twist make the same progress under both methods (0.67), so the gain lies in reaching and finishing stages rather than in progress within a stage (Table~\ref{tab:cond-pass}).

\paragraph{On Drill, the drilling stage itself is unchanged.}
Trials that reach drilling make similar progress (0.45 for SiMDex vs.\ 0.42 for GR-Dexter), and none completes it.
The gap arises earlier, at Grasp (4 vs.\ 6 completed) and Assembly (6 of 8 vs.\ 8 of 8).
SiMDex enters Assembly in 8 trials although it completes Grasp in only 4, because four of these entries use the trigger-press exception, against two for GR-Dexter.
This matches App.~\ref{app:mined-failures}, where no clip retrieved for the drilling stage shows drilling.

\paragraph{How to read these numbers.}
With ten trials per method, these counts are descriptive only. Table~\ref{tab:main}'s comparison is tested over layout-matched pairs in Sec.~\ref{sec:main_results} and App.~\ref{app:eval:aggregation}.

\subsection{Qualitative Rollouts}
\label{sec:supp-rollouts}

\begin{figure}[t]
\centering
\includegraphics[width=\textwidth]{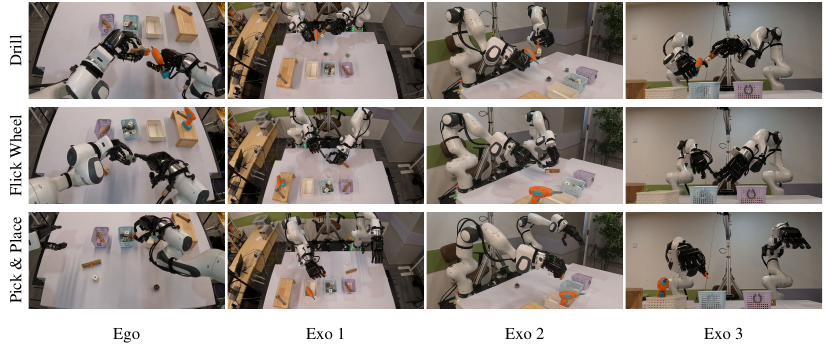}
\caption{\textbf{The four synchronized camera views.} One row per task: the same instant of one rollout from all four cameras. The exocentric cameras are used for recording only, never in training.}
\label{fig:fourview}
\end{figure}

Figure~\ref{fig:rollout-gallery} shows SiMDex rollouts on all ten sub-tasks, one row per sub-task.
The five columns of each row sample progress \emph{within} that sub-task, taken at even intervals from the start of the sub-task to its end, and labeled $t=0\%$ through $t=100\%$.
We sample by progress rather than by wall-clock time because the three tasks were recorded at different playback speeds, so sampling by time would make the rows incomparable across tasks.
All frames come from the Exo~2 side-on camera of Figure~\ref{fig:fourview}, the one view that shows both the state of the bench and the contact inside the hand.
Every frame in this subsection and the next is cropped only to remove the top and bottom bands carrying the burnt-in text overlay, with no zoom and no reframing of any kind.

These are single rollouts that illustrate what the sub-task columns of Table~\ref{tab:main} mean, not the ten-trial means: the Drill rollout shown does not complete the drilling sub-task.
The Flick Wheel row comes from an additional rollout recorded outside the scored trials, shown to illustrate the behavior rather than typical performance; in the scored trials of Table~\ref{tab:main}, that sub-task averages $0.10$.

\subsection{Multi-View Recording Setup}
\label{sec:supp-multiview}

Every rollout is recorded by four synchronized cameras: one egocentric view and three exocentric views.
Figure~\ref{fig:fourview} shows the same instant of the same rollout from all four cameras, one row per task.
The three exocentric cameras are used for recording only and were never used in training.
Scoring is done live by the evaluation team of App.~\ref{app:eval:independent}, not from any recording, so the multi-view setup does not affect any number in Table~\ref{tab:main}.
We keep this setup for post-hoc failure analysis, because different cameras expose different failure modes.
For Pick \& Place, Exo~3 sees the screw clear the hand's silhouette against a plain wall, while the top-down cameras see it only against cluttered trays.

\section{\texorpdfstring{Stage-Wise Ablation Details}{Stage-Wise Ablation Details}}
\label{app:stagewise}

This appendix details the stage-wise ablation of RQ3 at $1\times$ (Table~\ref{tab:stage-ablation}).

\paragraph{Setup.}
Each cascade run post-trains the same model on the cascade output up to a given stage, with the same training steps and 1:1 mixture as Table~\ref{tab:main}, so all runs see the same number of human samples.
Re-ranking (Stage III) only reorders and verifies the samples it receives: with the top 100 per anchor, as in Table~\ref{tab:main}, it would reorder the same 100 samples that Ranking (Stage II) returns, so the runs with and without it would train on identical data, leaving nothing to compare.
Each cascade run therefore keeps the top 50 per anchor: Ranking its own top 50, and Re-ranking its best 50 of the top 100 from Ranking.
Training budget, subset size and deduplication are thus the same across the cascade runs, where deduplication merges the per-anchor lists and keeps each sample once.
The cascade runs also mine from a subset of the anchors, whereas Table~\ref{tab:main} uses all anchors, so their numbers of distinct samples differ from that of Table~\ref{tab:main}.
All trials are scored by the stage-entry rule of App.~\ref{app:eval:scoring}.
Trials are paired by task and layout, and by round when both runs have two rounds (otherwise the two rounds are first averaged per layout).
Paired differences use unrounded trial scores, and $p$ is from a two-sided exact paired permutation test over these pairs.

\begin{table}[!ht]
\centering
\caption{\textbf{Other explanations of the gain from Ranking.} GR-Dexter* is the equal-size random subset of Table~\ref{tab:main}. The pooled difference is over the four robot-data budgets (App.~\ref{sec:supp-scaling-full}).}
\label{tab:app-alternatives}
\small
\begin{tabular}{@{}>{\raggedright\arraybackslash}p{2.6cm}>{\raggedright\arraybackslash}p{5.2cm}>{\raggedright\arraybackslash}p{5.2cm}@{}}
\toprule
Explanation & How it is tested & Result \\
\midrule
More distinct samples & GR-Dexter*, with more distinct samples than Ranking (1.49M vs.\ 1.05M) & 38.0 at $1\times$, vs.\ 58.3 for Ranking ($+$20.3) \\
\addlinespace[3pt]
Broad task relevance & Recall (Stage I) only, which selects by instruction and initial-pose similarity & 40.6, no better than GR-Dexter (47.7; $p = 0.44$) \\
\addlinespace[3pt]
Source diversity & GR-Dexter*, from twice as many episodes as SiMDex ($\sim$124,000 vs.\ $\sim$63,000; App.~\ref{app:mined-coverage}) & Below SiMDex at every budget ($+$17.8 pooled, $p < 0.0001$) \\
\bottomrule
\end{tabular}
\end{table}

\paragraph{Rounds per row.}
In Table~\ref{tab:stage-ablation}, Recall alone and the two Table~\ref{tab:main} rows average two rounds of five trials, and the other rows use one round.
The Table~\ref{tab:main} rows keep both rounds so that they match Table~\ref{tab:main}.
Recall alone has a second round because its first-round Drill score of zero looked surprising to us. The result held: in none of its ten Drill trials did it hold the drill stably. Both rounds are reported.
Unequal rounds favor no row. More trials make an estimate more precise, not higher or lower.
Every finding of Sec.~\ref{sec:stage-ablation} also holds with first rounds only: Recall alone scores 40.9, still no better than the first round of GR-Dexter (45.3, $p = 0.76$), and Ranking still adds $+17.4$ ($p = 0.015$).
One-round rows have no round-to-round spread, so Table~\ref{tab:stage-ablation} reports no $\pm$.

\paragraph{Other explanations of the gain.}
Table~\ref{tab:app-alternatives} tests three other explanations of the gain from Ranking.
The number of distinct samples alone is not enough: GR-Dexter* has more of them, yet scores far below Ranking.
Task relevance alone is not enough: Recall picks task-relevant clips, yet is no better than GR-Dexter.
Source diversity alone is not enough: GR-Dexter* draws on twice as many episodes, yet stays below SiMDex at every budget.
What all three controls lack is motion ranking, which supports the core design of SiMDex: selecting human data by hand motion.

\paragraph{Smaller mined sets.}
The two cascade runs with Ranking also give two smaller mined sets: 1.05M and 1.03M distinct samples, about 30\% fewer than the 1.49M of SiMDex.
Against SiMDex, their paired differences are $-3.3$ ($p = 0.69$) and $+0.2$ ($p = 0.99$), neither significant.
They also score above the larger GR-Dexter* at $1\times$ (38.0), by $+20.3$ and $+23.9$ points, respectively.

\end{document}